\documentclass{article}
\usepackage{ijcai20}

\usepackage{times}
\usepackage{soul}
\usepackage{url}
\usepackage[hidelinks]{hyperref}
\usepackage[utf8]{inputenc}
\usepackage[small]{caption}
\usepackage{graphicx}
\usepackage{amsmath}
\usepackage{booktabs}
\usepackage{algorithm}
\usepackage{algorithmic}
\usepackage{pdfpages}
\usepackage{pgfplots}
\usepackage{pgfplotstable}
\usepackage{color}
\usepackage{float}
\usepackage{booktabs}
\usepackage{multirow}
\usepackage{array}
\usepackage{amsfonts}
\usepackage{natbib}
\usepackage{mathtools}
\usepackage{subfig}
\usepackage{threeparttable}
\title{Generalized Embedding Machines for Recommender Systems}

\author{
Enneng Yang$^{1*}$
\and
Xin Xin$^{2}$\thanks{The co-first authors contribute equally} \and
Li Shen$^{3}$\And
Guibing Guo$^1$\footnote{Corresponding author}
\affiliations
$^1$Northeastern University, China\\
$^2$University of Glasgow, UK\\
$^3$Tencent AI Lab, China\\
\emails
1871149@stu.neu.edu.cn,
x.xin.1@research.gla.ac.uk,
mathshenli@gmail.com,\\
guogb@swc.neu.edu.cn
}

\begin{document}
\maketitle
\begin{abstract}
   Factorization machine (FM) is an effective model for feature-based recommendation which utilizes inner product to capture second-order feature interactions.
   However, one of the major drawbacks of FM is that it couldn't capture complex high-order interaction signals. A common solution is to change the interaction function, such as stacking deep neural networks on the top of FM. In this work, we propose an alternative approach to model high-order interaction signals in the embedding level, namely Generalized Embedding Machine (GEM). The embedding used in GEM encodes not only the information from the feature itself but also the information from other correlated features. Under such situation, the embedding becomes high-order. Then we can incorporate GEM with FM and even its advanced variants to perform feature interactions. More specifically, in this paper we utilize graph convolution networks (GCN) to generate high-order embeddings.
   We integrate GEM with several FM-based models and conduct extensive experiments on two real-world datasets. The results demonstrate significant improvement of GEM over corresponding baselines.
\end{abstract}

\section{Introduction}
Recommender systems has been widely deployed in practical applications to overcome the information overload issues, such as e-commerce websites, news portals, music apps, etc. It can benefit both users through providing the most interesting items and service providers by increasing traffic and profits \citep{adomavicius2005toward}. When training a recommender from user-item interactions, despite the most important IDs, rich side information has also become more common \citep{Xin2019CFM}. The side information contains but is not limited to user profile, item attributes, daytime of the transactions \citep{DBLP:conf/recsys/AdomaviciusT08}, etc. To represent the rich side information, a common solution is to convert them to sparse feature vectors through one/multi-hot encoding \citep{he2017neural}. After that, predictive models can be built based on the featured inputs.


To effectively learn recommenders from sparse feature vectors, it's important to model the interactions between features \citep{wang2017deepcross}. Conventional methods to capture feature interactions which are dependant on manually constructed cross features have been investigated a lot in both industry and academia \citep{cheng2016wide}. However, this kind of methods require heavy engineering efforts and domain knowledge, making it suffer from poor generalization ability and scalability.
Factorization machine (FM) \citep{rendle2010factorization} is another effective solution for feature-based recommendation. It first associates each feature with an embedding representation and then
automatically learns the feature interactions through inner product between corresponding embeddings. FM learns pair-wise (second-order) feature interactions in linear time complexity, achieving super success when handling sparse and high-dimensional feature vectors.

However, one of the major drawback of FM is that it can only model second-order feature interactions while more complex high-order information is actually very common in real-world use cases. For example, in movie recommendation scenario, it's hard to learn enough knowledge from the second-order interaction between $weekday=Saturday$ and $daytime=evening$. However, the high (third)-order interaction among $weekday=Saturday$, $daytime=evening$ and $age=8$ may imply that a cartoon movie could be a proper recommendation.
Based on such observation, some work has been proposed to capture high-order interaction signals. Existing solutions for high-order interaction modeling are based on either shadow methods such as kernel functions \citep{DBLP:conf/sdm/GuoAS18} and tensor decomposition \citep{DBLP:conf/wsdm/CaoZLY16, DBLP:conf/wsdm/LuHSCY17} or deep neural networks \citep{guo2017deepfm,he2017neural,Xin2019CFM}. However, all of these methods attempt to capture high-order signals by designing more complex interaction functions. On the other hand, the high-order signals can also be captured if the embedding itself is ``high-order''. This motivation conforms with the nature of cross features.


In this paper, we propose \emph{Generalized Embedding Machine} (GEM) to capture high-order interaction signals from the embedding level. The embedding used in GEM encodes the information from not only the feature itself but also the other correlated features. After that, even the simple inner product used in FM can also contains high-order signals because the feature embedding itself is high-order. Moreover, GEM can also be integrated with more advanced interaction functions used in the variants of FM, such as NFM \citep{he2017neural} and DeepFM \citep{guo2017deepfm}. Generally speaking, GEM can be seen as the combination of cross features and automatic interaction learning.
To make it more specific, in this paper we use graph convolution networks (GCN) \citep{Kipf2016GCN} to generate embeddings\footnote{GCN are not the only choice. We are also interested in exploiting other techniques (e.g., Transformer \citep{vaswani2017attention}).}. Each feature is represented as a graph node, and the edge can be defined according to the transactions of feature vectors.
Then GCN can perform information propagation and the output can be seen as high-order embeddings.

The main contributions of this work are summarized as follows:
\begin{itemize}
    \item We propose the GEM method which attempts to capture high-order interaction signals from the embedding level. It can be easily integrated with FM and its advanced variants.
    \item We propose to use GCN to construct high-order embeddings for GEM. It is worth mentioning that when we use one layer of GCN, GEM does not introduce any additional trainable parameters.
    \item We conduct extensive experiments on two datasets when integrating GEM with FM and its variants.
    Experimental results demonstrate the effectiveness of the GEM.
\end{itemize}

\section{Preliminaries}
\subsection{Factorization Machines}
In FM, the user-item interaction is represented by a transaction of sparse categorical feature vector $\mathbf{x} \in \mathbb{R}^m$. It utilizes one/multi-hot encoding to depict side information. An example is illustrated as follows.
\begin{equation*}
   \underbrace{[0,0,1,...,0]}_{\text{user ID}}\hspace{0.1cm}
   \underbrace{[0,1,0,...,0]}_{\text{item ID}}\hspace{0.1cm}
   \underbrace{[0,0,...,1]}_{\text{country=UK}}\hspace{0.1cm}
   \underbrace{[0,1,...,0]}_{\text{city=London}}\hspace{0.1cm}
\end{equation*}
The scoring function of FM is defined as the sum of a linear regression part and the second-order feature interactions:
\begin{equation}
    \label{original FM}
    \hat{y}_{FM}\left(\mathbf{x} \right) = {w}_{0} + \sum_{i=1}^{m}{w}_{i}{x}_{i} + \sum_{i=1}^{m}\sum_{j=i+1}^{m}{x}_{i}{x}_{j}\cdot\langle\mathbf{v}_i,\mathbf{v}_j\rangle,
\end{equation}
where ${w}_{0}$ represents the global bias, ${w}_{i}$ represents the bias for the $i$-th feature. The second-order interaction between feature $x_i$ and $x_{j}$ is captured by the inner product of their embeddings: $\langle\mathbf{v}_i,\mathbf{v}_j\rangle=\sum_{f=1}^{d}{v}_{if}{v}_{jf}$.
$\mathbf{v}_i\in \mathbb{R}^d$ can be seen as the embedding vector for feature $x_i$.

The time complexity of directly calculating Eq.(\ref{original FM}) is $\mathcal{O}(dm^2)$,  \cite{rendle2010factorization} reformulated the pair-wise interaction and reduced the time complexity to $\mathcal{O}(dm)$ through the following tricks:
\begin{equation}
    \label{reformulating FM}
    \resizebox{.91\linewidth}{!}{$
    \displaystyle
     \sum_{i=1}^{m}\sum_{j=i+1}^{m}{x}_{i}{x}_{j}\cdot\langle\mathbf{v}_i,\mathbf{v}_j\rangle
     = \frac{1}{2} \sum_{f=1}^d \Big( \big(\sum_{i=1}^m v_{i,f}x_i \big)^2 - \sum_{i=1}^m v^2_{i,f}x^2_i \Big)
$}
\end{equation}
It's obvious that FM only models the second-order interactions in a linear fashion. As discussed before, it's insufficient to learn complex high-order interaction signals in practical usage.
\subsection{Related Work}
FM \citep{rendle2010factorization} has achieved great success in both industry and academia due to it's effectiveness and high efficiency when handling sparse and high-dimensional feature vectors. Plenty of work has been done to improve FM, especially regarding to the modeling of high-order interactions.
HOFM \citep{DBLP:conf/nips/BlondelFUI16, DBLP:conf/icml/BlondelIFU16} and  SHFM \citep{DBLP:conf/sdm/GuoAS18} reformulate FM from the perspective of kernel function. The $k$-order ANOVA kernel \citep{stitson1999support} can be used to capture $k$-order feature interactions. MVM \citep{DBLP:conf/wsdm/CaoZLY16} and MFM \citep{DBLP:conf/wsdm/LuHSCY17} utilized the concept of multi-view learning to capture high-order interactions through tensor decomposition \citep{DBLP:journals/siamrev/KoldaB09}. However, this kind of methods are still limited to linear functions and cannot generalize to complex non-linear signals.

Recently, research about using deep neural networks \citep{lecun2015deep} to enhance FM has also attracted much attention. \cite{he2017neural} proposed NFM in which the second-order interaction are represented by a pooling vector. Then multi-layer perceptron (MLP) is stacked to extract high-order interaction signals. DeepFM \citep{guo2017deepfm} borrows the idea from Wide\&Deep \citep{cheng2016wide} and ensembles FM with another deep neural network. AFM \citep{AFM} proposed to use attention mechanism to assign different weights for different feature interactions. INN \citep{DBLP:conf/aaai/HongHC19} can be regarded as an upgraded version of AFM, which calculates two weights for both feature interactions and field interactions. IFM \citep{DBLP:conf/ijcai/YuWY19} proposed that the feature interactions in different transactions should have different weights and then utilized a memory network \citep{DBLP:journals/corr/WestonCB14} to learn the weights. \cite{Xin2019CFM} proposed to use outer-product and convolution neural networks \citep{ji20133dcnn} to model high-order interactions more explicitly.

We can see that all of these methods are actually working on designing more complex interaction functions to model high-order signals. However, another perspective is to capture the signal from the embedding level. Actually these two kinds of solutions can work together for better prediction, which is the proposition of this work.

\section{Methodology}
In this section, we first introduce the proposed GEM method when being integrated with FM. Then we illustrate a specific design of using GCN to generate high-order embeddings. We also provide a discussion to analysis the properties of GEM.
\subsection{Generalized Embedding Machines}
We can see from Eq.(\ref{original FM}) that FM models second-order interactions through the inner product between feature embeddings (i.e., $\langle\mathbf{v}_i, \mathbf{v}_j\rangle$). Here, the embedding $\mathbf{v}_i$ is generated through a simple embedding lookup operation. Its semantic meaning is just the representation of feature $x_i$. In the proposed GEM method, we expect this embedding to be ``high-order'', which means that it can reflect the information from not only the feature $x_i$ but also other correlated features. To achieve this, we can replace $\mathbf{v}_i$ with a more general and complex embedding function $\mathbf{g}(x_i) \in \mathbb{R}^d$. From this perspective, the scoring function of GEM when being integrated with FM can be formulated as follows:
\begin{equation}
    \label{original GEM}
    \hat{y}_{GEM}\left(\mathbf{x} \right) = {w}_{0} + \sum_{i=1}^{m}{w}_{i}{x}_{i} + \sum_{i=1}^{m}\sum_{j=i+1}^{m}{x}_{i}{x}_{j}\cdot\langle\mathbf{g}(x_i),\mathbf{g}(x_j)\rangle,
\end{equation}
If the introduced $\mathbf{g}(x_i)$ is capable of encoding high-order information, we can believe that even if we use simple inner product as the interaction function, the prediction of GEM (i.e., $\hat{y}_{GEM}$) still captures high-order information.


Similar to Eq.(\ref{reformulating FM}), we can also reformulate the interaction term as:
\begin{equation}
    \label{reformulating GEM}
    \frac{1}{2} \sum_{f=1}^d ( (\sum_{i=1}^m \mathbf{g}(x_i)_f x_i)^2 - \sum_{i=1}^m  \mathbf{g}(x_i)_f^2 x^2_i),
\end{equation}
where $\mathbf{g}(x_i)_f$ denotes the $f$-th element of $\mathbf{g}(x_i)$. It means that the proposed GEM would not change the original time complexity of the base model.

\subsection{GCN as the Embedding Function}
The motivation of GEM is the introduced $\mathbf{g}(x_i)$ can encode high-order information.
GCN is an effective representation learning approach based on graph data. It performs information propagation on the graph and the output embedding for a node is the aggregation of its neighbours. As a result, it provide a natural solution to generate high-order embeddings.

\begin{figure}[tb]
\centering
\includegraphics[width=0.42\textwidth]{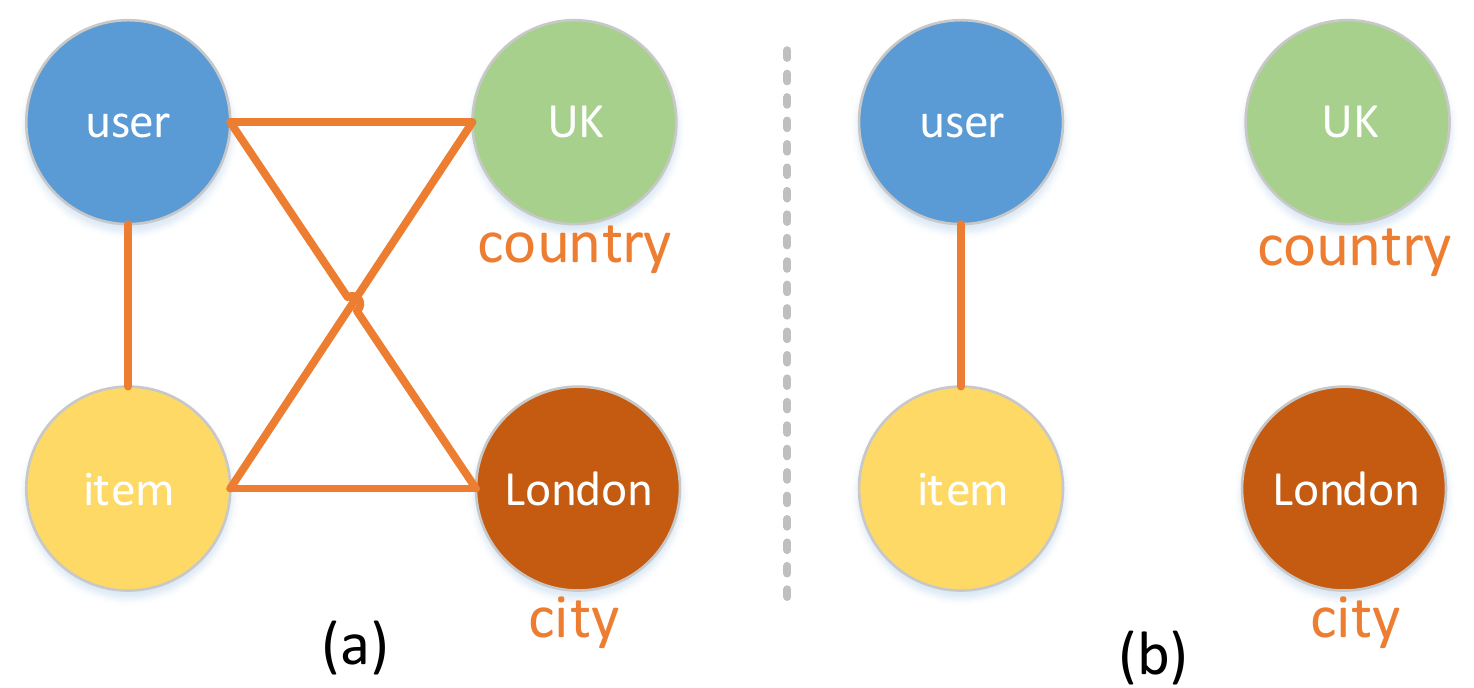}
\caption{Graph construction example on Frappe dataset}
\label{fig:gem-graph}
\end{figure}

From this perspective, we need to construct a graph based on the input feature vectors.
Generally speaking, we can represent each feature as a node in the graph. Then we can add an edge between two feature nodes if they occur in one feature vector.
Figure \ref{fig:gem-graph} demonstrates two examples when constructing the sub-graph for the illustrated feature vector in section 2.1.
In Figure \ref{fig:gem-graph}(a), the edge is defined among user-item, user-context (i.e., city and country) and item-context, while in Figure \ref{fig:gem-graph}(b), the edge is only defined among user-item.



\begin{figure}[tb]
\centering
\includegraphics[width=0.48\textwidth]{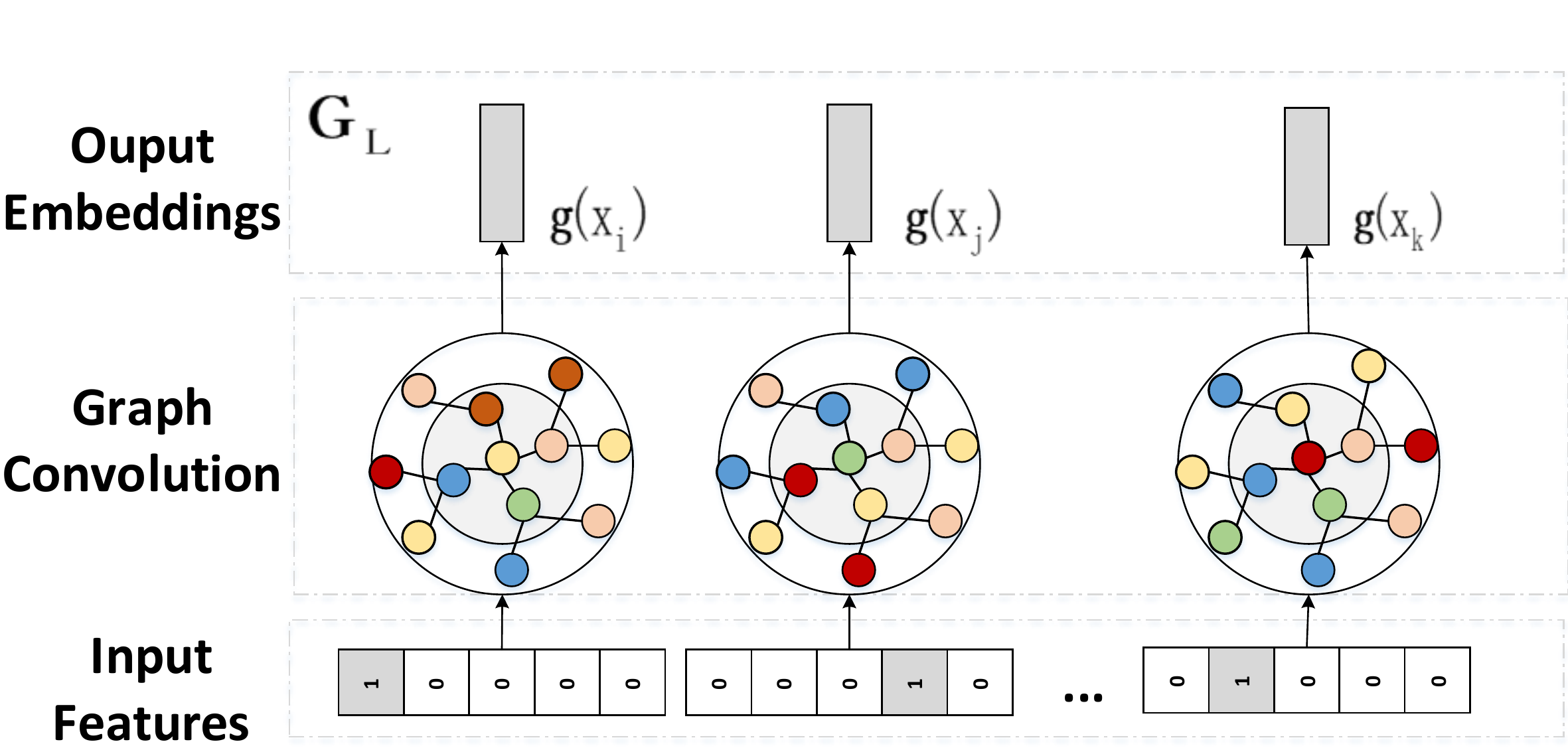}
\caption{Using GCN to generate high-order embeddings.}
\label{fig:gem-embedding}
\end{figure}

Given the constructed graph, let $\mathbf{A}$ denote the adjacent matrix of the graph, the whole embedding matrix $\mathbf{G}_1$ after one layer of graph convolution can be formulated as:
\begin{equation}
    \label{GCN whole embedding matrix}
    \mathbf{G}_1=\sigma(\widetilde{\mathbf{D}}^{-\frac{1}{2}}\widetilde{\mathbf{A}}\widetilde{\mathbf{D}}^{-\frac{1}{2}}\mathbf{W}_1\mathbf{H}_0),
\end{equation}
where $\widetilde{\mathbf{A}}=\mathbf{A+I}$, $\widetilde{\mathbf{D}}$ is the corresponding diagonal degree matrix of $\widetilde{\mathbf{A}}$ and $\sigma$ is the activation function.
$\mathbf{W}_1$ is the trainable parameter of the first graph convolution layer. $\mathbf{H}_0$ is the initial input which represents the properties contained in the node.
The information propagation can be repeated multi-times by replacing $\mathbf{H}_0$ with the output of the last convolution layer. We use $\mathbf{G}_L$ to denote the final output after $L$ layer of graph convolution.
After that, the embedding of feature $x_i$ is the $i$-th row of $\mathbf{G}_L$ as $\mathbf{g}(x_i)=\mathbf{G}_L(i,\cdot)$. Figure \ref{fig:gem-embedding} illustrates the procedure when using GCN to generate high-order embedding. We can see that if we use only one layer of GCN, the trainable parameters will be equals to $\mathbf{W}_1$, which has the same size with the embeddding table used in FM. In other words, we don't introduce additional parameters under such situation. Note that we can also try different variants of GCN, such as as GraphSAGE \citep{DBLP:conf/nips/HamiltonYL17} and graph attention network \citep{DBLP:conf/iclr/VelickovicCCRLB18}.


\subsection{Training Detail}
GEM can be used for regression, classification and ranking prediction tasks, just the corresponding optimization function. In this paper, the main focus is on the regression task.
\begin{equation}
    L = \sum_{\mathbf{x} \in \mathbb{S}} (\hat{y}_{\text{GEM}}(\mathbf{x}) - y(\mathbf{x}))^2 + \lambda ||\mathbf{\Theta}||^2.
    \label{eq:gem_loss}
\end{equation}
We add L2 regularization to avoid overfitting. $\lambda$ controls the regularization strength and $\mathbf{\Theta}$ denotes the parameters that need to be trained. We can use gradient descent to solve such optimization problems.


\subsection{Discussion}
Here we provide a brief discussion regarding the properties of GEM. Firstly, GEM can be easily downgraded to FM by setting $\mathbf{g}(x_i)=\mathbf{v}_i$. In that case, the embeddings contain no high-order information. Secondly, although we describe GEM by using FM as the base model, GEM can be also integrated with other advanced variants of FM by using $\mathbf{g}(x_i)$ as the input embedding. Finally, in this paper we use GCN as the embedding function. However, we can also use other techniques to generate high-order embeddings. For example, we can treat each feature as a word and the input feature vector as a sentence. Then we can utilize the Transformer encoder \citep{vaswani2017attention} as the embedding function. The hidden output of the encoder can be seen as the generated high-order embeddings.

\section{Experiments}
In this section, we conduct experiments on two real-world datasets to evaluate the effectiveness of the proposed GEM method.
We intend to answer the following research questions:
\begin{itemize}
    \item \textbf{RQ1} Does GEM help to improve the performance when integrated with different base models?
    \item \textbf{RQ2} How does the design of GCN affect the performance when we use it as the embedding function?
    \item \textbf{RQ3} How do the key hyperparameter (the dropout ratio) of GEM impact its performance?
\end{itemize}

\subsection{Experimental Settings}


\subsubsection{Data Description}
We evaluate our models on two publicly datasets: Frappe\footnote{http://baltrunas.info/research-menu/frappe} and MovieLens\footnote{https://grouplens.org/datasets/movielens/latest/}.

\textbf{Frappe.} The Frappe dataset is a context-aware app recommendation dataset conducted by \cite{baltrunas2015frappe}. The original Frappe contains 96,203 records of 957 user's clicks on 96203 apps. Each transaction contains user ID, app ID, and eight context features. 
We noticed that some features have very few nodes in the graph. For example, the feature $isweekend$ only has two nodes. This results into a situation that the nodes of these features are actually connected to almost all the other nodes in the graph and thus introduce noise into the learning of GCN. So we only take the features describing countries and cities as the nodes in the graph. We follow the pre-processing method of \cite{he2017neural}, and take two apps that have not been clicked under the same context as negative samples, resulting into a dataset contains 288,609 instances.

\textbf{MovieLens.} The MovieLens dataset is published by GroupLens \citep{DBLP:journals/tiis/HarperK16}. We follow the same procedure of \citep{he2017neural,DBLP:conf/ijcai/YuWY19} to process the dataset as personalized tag recommendation. The data obtained included 668,953 records of 17,045 users on 23,743 items of 49,657 tags. The total 2,066,859 instances of data can be obtained by using the same negative sampling technique with Frappe.

\subsubsection{Evaluation Protocols}
We use the same data split with \citep{DBLP:conf/ijcai/YuWY19} in which the ratio of training, validation and test set as 8: 1: 1. In this paper, our task is rating prediction so we chose two most used  indicators in regression tasks as the evaluation metrics: Mean Absolute Error (MAE) and Root Mean Squard Error (RMSE). Lower MAE and RMSE values indicate better prediction accuracy.
\subsubsection{Baselines}
All models are implemented by TensorFlow. We use the author's official implementation if it is public accessible. We integrate GEM with the following baseline:
\begin{itemize}
    \item FM \citep{rendle2010factorization}: This is the original FM. We rewrote the official implementation\footnote{http://www.libfm.org/} \citep{rendle2012factorization} using TensorFlow and introduced dropout after feature interactions to improve the performance.
    \item AFM \citep{AFM}: It uses the attention network to learn the weight of feature interaction. We refer to the original paper and use one layer of attention network. The hidden size of the attention network is tuned between $\left\{16, 32, 64, 128, 256 \right\}$.
    \item Wide$\&$Deep \citep{cheng2016wide}: This is composed of two parts: linear regression as wide part and MLP as deep part. For the deep part, we use three layer MLP, and the number of neurons corresponding to each layer is set as 1024, 512 and 256, respectively.
    \item DeepFM \citep{guo2017deepfm}: It is similar to Wide$\&$Deep. The wide part is replaced by FM. For the deep part, we refer to the original paper and utilize three-layer MLP with 200 neurons in each layer.
    \item NFM \citep{he2017neural}: It models high-order interaction signals by stacking MLP upon the pooling interaction vectors. We refer to the original paper and use one layer of MLP. The number of neurons is searched in $\left\{64, 128, 256\right\}$.
    \item INN \citep{DBLP:conf/aaai/HongHC19}: In addition to considering the importance of feature interactions, it also learns the importance of field interactions. Similar to AFM, the size of the attention network for feature interaction is searched from $\left\{16, 32, 64, 128, 256 \right\}$.  For the field interaction, the hidden size is tuned between $\left\{8, 16, 24, 32 \right\}$.
    \item IFM \citep{DBLP:conf/ijcai/YuWY19}: It uses a memory network to learn the weight of the  feature in each transaction. We set the number of neurons in the memory network according to the original paper.
\end{itemize}

\subsubsection{Other Parameter Settings}
For the sake of fair comparison, all models are run in the same environment. The embedding size is set as 256 for all models on both datasets.
We train all models using square loss.
The batch size is set to 4096. We noticed that models achieve different results when using different optimizers and we selected the best results between Adagrad \citep{DBLP:journals/jmlr/DuchiHS11} and Adam \citep{DBLP:journals/corr/KingmaB14,zou2019sufficient} respectively. According to our experiment, AFM, IFM, INN works better when using Adagrad, while other models performs better when using Adam. Learning rate was searched between $\left\{0.001, 0.002, 0.005, 0.01\right\}$. L2 regularization coefficient was tuned from   $\left\{1e-3,1e-4,1e-5\right\}$ and the dropout \citep{DBLP:journals/jmlr/SrivastavaHKSS14} ratio was searched in  $\left\{0.0,0.1,\ldots,0.9\right\}$. For GEM, we use the method illustrated in Figure \ref{fig:gem-graph}(a) to construct the graph without special mention.

For AFM, Wide$\&$Deep, DeepFM, NFM, IFM and INN, since the performance after pre-training is better, we use the embeddings of FM as pre-trained parameters. The original FM and all our GEM methods don't use pre-training. In all models, we use the early-stop strategy \citep{zhang2005boosting} and end the training when the result doesn't improve on the validation set for 5 consecutive times.
\subsection{Performance Comparison RQ1)}
In this section, we compare the performance of GEM when integrated with different base models. We also investigate the convergence of the GEM when using FM as the base model.

\begin{table*}[tb]
\centering
\begin{tabular}{|l|l|l|l|l|l|l|}
\hline
\multicolumn{1}{|c|}{\multirow{2}{*}{\textbf{Method}}} & \multicolumn{3}{c|}{\textbf{Frappe}}            & \multicolumn{3}{c|}{\textbf{Movielens}}         \\ \cline{2-7}
\multicolumn{1}{|c|}{}                        & \textbf{\#Param} & \textbf{MAE} & \textbf{RMSE} & \textbf{\#Param} & \textbf{MAE} & \textbf{RMSE} \\
\hline \hline
FM
&1.383M   &0.1445   &0.3143
&23.24M    &0.2425      &0.4431           \\
\textbf{GEM+FM}
&1.383M   &\textbf{0.1188}   &\textbf{0.3027}
&23.24M    &\textbf{0.2243} &\textbf{0.4307}        \\ \hline \hline
AFM
&1.449M   &0.1459   &0.3131
&23.26M    &0.2656      &0.4506           \\
\textbf{GEM+AFM}
&1.449M   &\textbf{0.1179}  &\textbf{0.3017}
&23.26M    &\textbf{0.2350}   &\textbf{0.4345}         \\
\hline \hline
Wide$\&$Deep
&4.662M   &0.0678         &\textbf{0.2981}
&24.69M    &0.1297      &\textbf{0.4212}          \\
\textbf{GEM+Wide$\&$Deep}
&4.662M   &\textbf{0.0611}         &0.3079
&24.69M    &\textbf{0.1108}      &0.4260        \\
\hline \hline
DeepFM
&1.975M   &0.0626   &0.2969
&23.47M    &0.1296      &0.4328          \\
\textbf{GEM+DeepFM}
&1.975M   &\textbf{0.0602}       &\textbf{0.2951}
&23.47M    &\textbf{0.1106}      &\textbf{0.4231}        \\
\hline \hline
NFM
&1.399M   &0.0733   &0.2921
&23.31M    &\textbf{0.1511}  &0.4285      \\
\textbf{GEM+NFM}
&1.399M   &\textbf{0.0710}   &\textbf{0.2918}
&23.31M    &0.1680      &\textbf{0.4200}         \\
\hline \hline
INN
&1.457M   &0.0701   &\textbf{0.2971}
&23.25M    &0.1745      &0.4096         \\
\textbf{GEM+INN}
&1.457M   &\textbf{0.0609}       &0.2976
&23.25M    &\textbf{0.1397}      &\textbf{0.4015}            \\
\hline \hline
IFM
&2.109M    &0.0565        &0.2905
&23.80M    &0.1561      &0.4252            \\
\textbf{GEM+IFM}
&2.109M    &\textbf{0.0557}   & \textbf{0.2890}
&23.80M    &\textbf{0.1411}      &\textbf{0.4156} \\
\hline
\end{tabular}
\centering
\caption{Performance comparison on Frappe and Movielens (embedding size: 256). Boldface denotes better results.}
\label{tab:performance}
\end{table*}

\subsubsection{Performance Analysis}
The performance of all models on Frappe dataset and MovieLens dataset is shown in Table \ref{tab:performance}. For the proposed GEM, we use one layer of GCN and the hidden size is also set as 256, which means GEM doesn't introduce any extra trainable parameters.
\begin{itemize}
    \item Compared with FM: We observed that GEM consistently outperformed the original FM in both evaluation metrics on the two datasets, indicating that capturing higher-order signals at the embedding level is effective to improve the model performance. The relative improvement of the two evaluation metrics can be calculated as follows: On Frappe dataset, MAE and RMSE are improved by $17.78 \%$ and $3.69 \%$, respectively.  On MovieLens dataset, MAE and RMSE are improved by $7.50 \%$ and $2.79\%$, respectively. Besides, we can also find that simple incorporating GEM into FM can even help FM to beat some advanced baselines. For example, the RMSE of GEM+FM is 0.3027 which is smaller than the result of AFM (i.e., 0.3131). It further confirms the effectiveness of our methods because we get better results with same or even smaller number of parameters.


    \item Compared with the variants of FM: For AFM, IFM and INN, the three models are not designed for modeling high-order interaction signals. When we use our GEM to introduce high-order information from the embedding level, most of the models are improved. It indicates that it is feasible to combine the proposed GEM with the advantages of other models, and almost all of them could further benefit from GEM.
    Wide$\&$Deep, DeepFM and NFM are proposed to capture high-order interaction signals by designing more complex interaction functions (i.e., staking deep neural networks). However, the performance still get improved when integrated with our GEM method. It demonstrate that considering high-order information from embedding level can work collaboratively with complex interaction functions.

\end{itemize}

\subsubsection{Convergence Behaviors Analysis}
Figure \ref{fig:comparison_rmse} and Figure \ref{fig:comparison_mae} show the convergence of FM and GEM in terms of RMSE and MAE, respectively. We can see that the final performance of GEM is better than FM on both datasets.
Besides, we can also see that GEM achieves faster and more stable convergence in the MovieLens dataset.
On the Frappe dataset, GEM fluctuates more than FM during the training process but the final performance is better. The reason of fluctuation may be that the Frappe dataset is relatively smaller.

\pgfplotsset{width=4.4cm, height=4.0cm, compat=1.5}
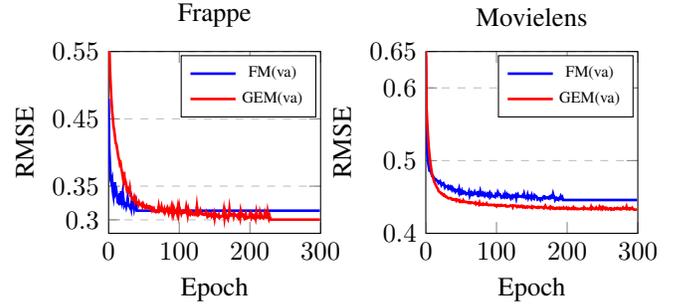
\begin{figure}[tb]
\begin{tikzpicture}
\begin{axis}[
    title={Frappe},
    xlabel={Epoch},
    ylabel={RMSE},
    xmin=0, xmax=300,
    ymin=0.28, ymax=0.55,
    xtick={0,100,200,300},
    ytick={0.30,0.35,0.45,0.55},
    ymajorgrids=true,
    xmajorgrids=false,
    grid style=dashed,
    legend entries={FM(va), GEM(va)},
    legend style={
            font=\tiny,
            /tikz/every even column/.append style={column sep=0.5cm}
        }
]
\addplot[
    color=blue,
      line width=1pt,
    ]
    coordinates {
    (0,0.4798)(1,0.4060)(2,0.3815)(3,0.3596)(4,0.3554)(5,0.3596)(6,0.3424)(7,0.3353)(8,0.3468)(9,0.3554)(10,0.3403)(11,0.3429)(12,0.3347)(13,0.3260)(14,0.3249)(15,0.3300)(16,0.3394)(17,0.3219)(18,0.3426)(19,0.3320)(20,0.3402)(21,0.3308)(22,0.3218)(23,0.3244)(24,0.3238)(25,0.3192)(26,0.3236)(27,0.3319)(28,0.3221)(29,0.3218)(30,0.3191)(31,0.3211)(32,0.3165)(33,0.3283)(34,0.3221)(35,0.3235)(36,0.3170)(37,0.3234)(38,0.3199)(39,0.3246)(40,0.3136)(41,0.3136)(42,0.3136)(43,0.3136)(44,0.3136)(45,0.3136)(46,0.3136)(47,0.3136)(48,0.3136)(49,0.3136)(50,0.3136)(51,0.3136)(52,0.3136)(53,0.3136)(54,0.3136)(55,0.3136)(56,0.3136)(57,0.3136)(58,0.3136)(59,0.3136)(60,0.3136)(61,0.3136)(62,0.3136)(63,0.3136)(64,0.3136)(65,0.3136)(66,0.3136)(67,0.3136)(68,0.3136)(69,0.3136)(70,0.3136)(71,0.3136)(72,0.3136)(73,0.3136)(74,0.3136)(75,0.3136)(76,0.3136)(77,0.3136)(78,0.3136)(79,0.3136)(80,0.3136)(81,0.3136)(82,0.3136)(83,0.3136)(84,0.3136)(85,0.3136)(86,0.3136)(87,0.3136)(88,0.3136)(89,0.3136)(90,0.3136)(91,0.3136)(92,0.3136)(93,0.3136)(94,0.3136)(95,0.3136)(96,0.3136)(97,0.3136)(98,0.3136)(99,0.3136)(100,0.3136)(101,0.3136)(102,0.3136)(103,0.3136)(104,0.3136)(105,0.3136)(106,0.3136)(107,0.3136)(108,0.3136)(109,0.3136)(110,0.3136)(111,0.3136)(112,0.3136)(113,0.3136)(114,0.3136)(115,0.3136)(116,0.3136)(117,0.3136)(118,0.3136)(119,0.3136)(120,0.3136)(121,0.3136)(122,0.3136)(123,0.3136)(124,0.3136)(125,0.3136)(126,0.3136)(127,0.3136)(128,0.3136)(129,0.3136)(130,0.3136)(131,0.3136)(132,0.3136)(133,0.3136)(134,0.3136)(135,0.3136)(136,0.3136)(137,0.3136)(138,0.3136)(139,0.3136)(140,0.3136)(141,0.3136)(142,0.3136)(143,0.3136)(144,0.3136)(145,0.3136)(146,0.3136)(147,0.3136)(148,0.3136)(149,0.3136)(150,0.3136)(151,0.3136)(152,0.3136)(153,0.3136)(154,0.3136)(155,0.3136)(156,0.3136)(157,0.3136)(158,0.3136)(159,0.3136)(160,0.3136)(161,0.3136)(162,0.3136)(163,0.3136)(164,0.3136)(165,0.3136)(166,0.3136)(167,0.3136)(168,0.3136)(169,0.3136)(170,0.3136)(171,0.3136)(172,0.3136)(173,0.3136)(174,0.3136)(175,0.3136)(176,0.3136)(177,0.3136)(178,0.3136)(179,0.3136)(180,0.3136)(181,0.3136)(182,0.3136)(183,0.3136)(184,0.3136)(185,0.3136)(186,0.3136)(187,0.3136)(188,0.3136)(189,0.3136)(190,0.3136)(191,0.3136)(192,0.3136)(193,0.3136)(194,0.3136)(195,0.3136)(196,0.3136)(197,0.3136)(198,0.3136)(199,0.3136)(200,0.3136)(201,0.3136)(202,0.3136)(203,0.3136)(204,0.3136)(205,0.3136)(206,0.3136)(207,0.3136)(208,0.3136)(209,0.3136)(210,0.3136)(211,0.3136)(212,0.3136)(213,0.3136)(214,0.3136)(215,0.3136)(216,0.3136)(217,0.3136)(218,0.3136)(219,0.3136)(220,0.3136)(221,0.3136)(222,0.3136)(223,0.3136)(224,0.3136)(225,0.3136)(226,0.3136)(227,0.3136)(228,0.3136)(229,0.3136)(230,0.3136)(231,0.3136)(232,0.3136)(233,0.3136)(234,0.3136)(235,0.3136)(236,0.3136)(237,0.3136)(238,0.3136)(239,0.3136)(240,0.3136)(241,0.3136)(242,0.3136)(243,0.3136)(244,0.3136)(245,0.3136)(246,0.3136)(247,0.3136)(248,0.3136)(249,0.3136)(250,0.3136)(251,0.3136)(252,0.3136)(253,0.3136)(254,0.3136)(255,0.3136)(256,0.3136)(257,0.3136)(258,0.3136)(259,0.3136)(260,0.3136)(261,0.3136)(262,0.3136)(263,0.3136)(264,0.3136)(265,0.3136)(266,0.3136)(267,0.3136)(268,0.3136)(269,0.3136)(270,0.3136)(271,0.3136)(272,0.3136)(273,0.3136)(274,0.3136)(275,0.3136)(276,0.3136)(277,0.3136)(278,0.3136)(279,0.3136)(280,0.3136)(281,0.3136)(282,0.3136)(283,0.3136)(284,0.3136)(285,0.3136)(286,0.3136)(287,0.3136)(288,0.3136)(289,0.3136)(290,0.3136)(291,0.3136)(292,0.3136)(293,0.3136)(294,0.3136)(295,0.3136)(296,0.3136)(297,0.3136)(298,0.3136)(299,0.3136)
    }; 

\addplot[
    color=red,
      line width=1pt,
    ]
    coordinates {
     (0,0.6176)(1,0.5630)(2,0.5239)(3,0.4999)(4,0.4788)(5,0.4616)(6,0.4512)(7,0.4387)(8,0.4326)(9,0.4239)(10,0.4141)(11,0.4125)(12,0.4063)(13,0.3992)(14,0.3924)(15,0.3892)(16,0.3824)(17,0.3841)(18,0.3765)(19,0.3711)(20,0.3650)(21,0.3589)(22,0.3711)(23,0.3556)(24,0.3570)(25,0.3517)(26,0.3440)(27,0.3456)(28,0.3410)(29,0.3408)(30,0.3467)(31,0.3359)(32,0.3344)(33,0.3325)(34,0.3295)(35,0.3297)(36,0.3333)(37,0.3277)(38,0.3313)(39,0.3273)(40,0.3256)(41,0.3242)(42,0.3220)(43,0.3264)(44,0.3242)(45,0.3228)(46,0.3219)(47,0.3235)(48,0.3200)(49,0.3233)(50,0.3205)(51,0.3206)(52,0.3174)(53,0.3175)(54,0.3206)(55,0.3185)(56,0.3179)(57,0.3200)(58,0.3203)(59,0.3169)(60,0.3188)(61,0.3188)(62,0.3138)(63,0.3153)(64,0.3168)(65,0.3126)(66,0.3126)(67,0.3116)(68,0.3129)(69,0.3120)(70,0.3121)(71,0.3197)(72,0.3136)(73,0.3154)(74,0.3143)(75,0.3191)(76,0.3136)(77,0.3133)(78,0.3118)(79,0.3156)(80,0.3111)(81,0.3138)(82,0.3149)(83,0.3176)(84,0.3136)(85,0.3100)(86,0.3144)(87,0.3109)(88,0.3113)(89,0.3174)(90,0.3102)(91,0.3169)(92,0.3097)(93,0.3150)(94,0.3114)(95,0.3131)(96,0.3145)(97,0.3098)(98,0.3175)(99,0.3096)(100,0.3185)(101,0.3117)(102,0.3119)(103,0.3088)(104,0.3120)(105,0.3131)(106,0.3106)(107,0.3110)(108,0.3109)(109,0.3127)(110,0.3197)(111,0.3146)(112,0.3074)(113,0.3143)(114,0.3092)(115,0.3080)(116,0.3127)(117,0.3133)(118,0.3085)(119,0.3078)(120,0.3096)(121,0.3100)(122,0.3083)(123,0.3106)(124,0.3091)(125,0.3198)(126,0.3106)(127,0.3071)(128,0.3081)(129,0.3097)(130,0.3139)(131,0.3110)(132,0.3087)(133,0.3091)(134,0.3079)(135,0.3087)(136,0.3076)(137,0.3079)(138,0.3111)(139,0.3077)(140,0.3043)(141,0.3052)(142,0.3086)(143,0.3070)(144,0.3091)(145,0.3096)(146,0.3053)(147,0.3067)(148,0.3050)(149,0.3092)(150,0.3024)(151,0.3095)(152,0.3085)(153,0.3067)(154,0.3047)(155,0.3045)(156,0.3073)(157,0.3032)(158,0.3044)(159,0.3053)(160,0.3051)(161,0.3062)(162,0.3066)(163,0.3138)(164,0.3034)(165,0.3114)(166,0.3067)(167,0.3040)(168,0.3049)(169,0.3019)(170,0.3074)(171,0.3164)(172,0.3075)(173,0.3064)(174,0.3076)(175,0.3082)(176,0.3090)(177,0.3047)(178,0.3093)(179,0.3083)(180,0.3066)(181,0.3066)(182,0.3050)(183,0.3072)(184,0.3041)(185,0.3069)(186,0.3026)(187,0.3130)(188,0.3083)(189,0.3064)(190,0.3074)(191,0.3047)(192,0.3027)(193,0.3067)(194,0.3093)(195,0.3020)(196,0.3025)(197,0.3034)(198,0.3032)(199,0.3046)(200,0.3097)(201,0.3109)(202,0.3038)(203,0.3064)(204,0.3037)(205,0.3030)(206,0.3072)(207,0.3069)(208,0.3026)(209,0.3104)(210,0.3035)(211,0.3040)(212,0.3040)(213,0.3034)(214,0.3104)(215,0.3053)(216,0.3074)(217,0.3032)(218,0.3034)(219,0.3024)(220,0.3050)(221,0.3100)(222,0.3055)(223,0.3031)(224,0.3037)(225,0.3122)(226,0.3059)(227,0.3015)(228,0.3049)(229,0.3003)(230,0.3003)(231,0.3003)(232,0.3003)(233,0.3003)(234,0.3003)(235,0.3003)(236,0.3003)(237,0.3003)(238,0.3003)(239,0.3003)(240,0.3003)(241,0.3003)(242,0.3003)(243,0.3003)(244,0.3003)(245,0.3003)(246,0.3003)(247,0.3003)(248,0.3003)(249,0.3003)(250,0.3003)(251,0.3003)(252,0.3003)(253,0.3003)(254,0.3003)(255,0.3003)(256,0.3003)(257,0.3003)(258,0.3003)(259,0.3003)(260,0.3003)(261,0.3003)(262,0.3003)(263,0.3003)(264,0.3003)(265,0.3003)(266,0.3003)(267,0.3003)(268,0.3003)(269,0.3003)(270,0.3003)(271,0.3003)(272,0.3003)(273,0.3003)(274,0.3003)(275,0.3003)(276,0.3003)(277,0.3003)(278,0.3003)(279,0.3003)(280,0.3003)(281,0.3003)(282,0.3003)(283,0.3003)(284,0.3003)(285,0.3003)(286,0.3003)(287,0.3003)(288,0.3003)(289,0.3003)(290,0.3003)(291,0.3003)(292,0.3003)(293,0.3003)(294,0.3003)(295,0.3003)(296,0.3003)(297,0.3003)(298,0.3003)(299,0.3003)
    };  
\end{axis}

\hskip 120pt
\begin{axis}[
    title={Movielens},
    xlabel={Epoch},
    ylabel={RMSE},
    xmin=0, xmax=300,
    ymin=0.4, ymax=0.65,
    xtick={0,100,200,300},
    ytick={0.4,0.5,0.6,0.65},
    ymajorgrids=true,
    xmajorgrids=false,
    grid style=dashed,
    legend entries={ FM(va),  GEM(va)},
    legend style={
            font=\tiny,
            /tikz/every even column/.append style={column sep=0.5cm}
        }
]

\addplot[
    color=blue,
      line width=1pt,
    ]
    coordinates {
  (0,0.6444)(1,0.5351)(2,0.5101)(3,0.4992)(4,0.4924)(5,0.4865)(6,0.4865)(7,0.4851)(8,0.4819)(9,0.4789)(10,0.4789)(11,0.4766)(12,0.4758)(13,0.4748)(14,0.4731)(15,0.4734)(16,0.4727)(17,0.4712)(18,0.4693)(19,0.4706)(20,0.4716)(21,0.4696)(22,0.4665)(23,0.4682)(24,0.4675)(25,0.4686)(26,0.4666)(27,0.4632)(28,0.4649)(29,0.4642)(30,0.4641)(31,0.4649)(32,0.4611)(33,0.4617)(34,0.4637)(35,0.4619)(36,0.4633)(37,0.4603)(38,0.4625)(39,0.4611)(40,0.4588)(41,0.4596)(42,0.4574)(43,0.4592)(44,0.4597)(45,0.4599)(46,0.4575)(47,0.4610)(48,0.4554)(49,0.4578)(50,0.4558)(51,0.4583)(52,0.4549)(53,0.4548)(54,0.4573)(55,0.4570)(56,0.4560)(57,0.4539)(58,0.4547)(59,0.4541)(60,0.4544)(61,0.4571)(62,0.4521)(63,0.4553)(64,0.4564)(65,0.4537)(66,0.4550)(67,0.4535)(68,0.4550)(69,0.4570)(70,0.4542)(71,0.4544)(72,0.4554)(73,0.4533)(74,0.4550)(75,0.4522)(76,0.4534)(77,0.4577)(78,0.4536)(79,0.4550)(80,0.4532)(81,0.4552)(82,0.4532)(83,0.4516)(84,0.4541)(85,0.4522)(86,0.4521)(87,0.4540)(88,0.4519)(89,0.4540)(90,0.4555)(91,0.4495)(92,0.4506)(93,0.4535)(94,0.4517)(95,0.4508)(96,0.4508)(97,0.4512)(98,0.4502)(99,0.4537)(100,0.4525)(101,0.4517)(102,0.4564)(103,0.4544)(104,0.4532)(105,0.4535)(106,0.4510)(107,0.4491)(108,0.4499)(109,0.4497)(110,0.4524)(111,0.4506)(112,0.4539)(113,0.4499)(114,0.4494)(115,0.4530)(116,0.4510)(117,0.4530)(118,0.4526)(119,0.4499)(120,0.4482)(121,0.4511)(122,0.4520)(123,0.4492)(124,0.4509)(125,0.4518)(126,0.4508)(127,0.4524)(128,0.4481)(129,0.4528)(130,0.4539)(131,0.4504)(132,0.4505)(133,0.4508)(134,0.4506)(135,0.4508)(136,0.4502)(137,0.4520)(138,0.4509)(139,0.4543)(140,0.4502)(141,0.4503)(142,0.4485)(143,0.4496)(144,0.4530)(145,0.4489)(146,0.4514)(147,0.4516)(148,0.4491)(149,0.4476)(150,0.4518)(151,0.4488)(152,0.4481)(153,0.4474)(154,0.4522)(155,0.4510)(156,0.4510)(157,0.4499)(158,0.4465)(159,0.4471)(160,0.4479)(161,0.4473)(162,0.4458)(163,0.4483)(164,0.4483)(165,0.4471)(166,0.4491)(167,0.4509)(168,0.4507)(169,0.4480)(170,0.4483)(171,0.4492)(172,0.4494)(173,0.4482)(174,0.4477)(175,0.4504)(176,0.4510)(177,0.4492)(178,0.4509)(179,0.4500)(180,0.4541)(181,0.4503)(182,0.4500)(183,0.4459)(184,0.4475)(185,0.4485)(186,0.4469)(187,0.4476)(188,0.4464)(189,0.4483)(190,0.4477)(191,0.4509)(192,0.4489)(193,0.4499)(194,0.4458)(195,0.4458)(196,0.4458)(197,0.4458)(198,0.4458)(199,0.4458)(200,0.4458)(201,0.4458)(202,0.4458)(203,0.4458)(204,0.4458)(205,0.4458)(206,0.4458)(207,0.4458)(208,0.4458)(209,0.4458)(210,0.4458)(211,0.4458)(212,0.4458)(213,0.4458)(214,0.4458)(215,0.4458)(216,0.4458)(217,0.4458)(218,0.4458)(219,0.4458)(220,0.4458)(221,0.4458)(222,0.4458)(223,0.4458)(224,0.4458)(225,0.4458)(226,0.4458)(227,0.4458)(228,0.4458)(229,0.4458)(230,0.4458)(231,0.4458)(232,0.4458)(233,0.4458)(234,0.4458)(235,0.4458)(236,0.4458)(237,0.4458)(238,0.4458)(239,0.4458)(240,0.4458)(241,0.4458)(242,0.4458)(243,0.4458)(244,0.4458)(245,0.4458)(246,0.4458)(247,0.4458)(248,0.4458)(249,0.4458)(250,0.4458)(251,0.4458)(252,0.4458)(253,0.4458)(254,0.4458)(255,0.4458)(256,0.4458)(257,0.4458)(258,0.4458)(259,0.4458)(260,0.4458)(261,0.4458)(262,0.4458)(263,0.4458)(264,0.4458)(265,0.4458)(266,0.4458)(267,0.4458)(268,0.4458)(269,0.4458)(270,0.4458)(271,0.4458)(272,0.4458)(273,0.4458)(274,0.4458)(275,0.4458)(276,0.4458)(277,0.4458)(278,0.4458)(279,0.4458)(280,0.4458)(281,0.4458)(282,0.4458)(283,0.4458)(284,0.4458)(285,0.4458)(286,0.4458)(287,0.4458)(288,0.4458)(289,0.4458)(290,0.4458)(291,0.4458)(292,0.4458)(293,0.4458)(294,0.4458)(295,0.4458)(296,0.4458)(297,0.4458)(298,0.4458)(299,0.4458)
    }; 

\addplot[
    color=red,
      line width=1pt,
    ]
    coordinates {
  (0,0.6821)(1,0.5735)(2,0.5515)(3,0.5347)(4,0.5198)(5,0.5094)(6,0.5002)(7,0.4928)(8,0.4865)(9,0.4802)(10,0.4777)(11,0.4739)(12,0.4704)(13,0.4685)(14,0.4667)(15,0.4644)(16,0.4625)(17,0.4613)(18,0.4581)(19,0.4572)(20,0.4565)(21,0.4559)(22,0.4543)(23,0.4546)(24,0.4534)(25,0.4522)(26,0.4530)(27,0.4516)(28,0.4503)(29,0.4505)(30,0.4494)(31,0.4488)(32,0.4480)(33,0.4482)(34,0.4484)(35,0.4486)(36,0.4478)(37,0.4465)(38,0.4469)(39,0.4463)(40,0.4460)(41,0.4467)(42,0.4469)(43,0.4449)(44,0.4465)(45,0.4463)(46,0.4451)(47,0.4440)(48,0.4452)(49,0.4434)(50,0.4440)(51,0.4432)(52,0.4432)(53,0.4453)(54,0.4435)(55,0.4431)(56,0.4433)(57,0.4430)(58,0.4431)(59,0.4428)(60,0.4418)(61,0.4422)(62,0.4416)(63,0.4424)(64,0.4417)(65,0.4422)(66,0.4423)(67,0.4418)(68,0.4415)(69,0.4411)(70,0.4422)(71,0.4402)(72,0.4415)(73,0.4410)(74,0.4413)(75,0.4398)(76,0.4424)(77,0.4397)(78,0.4396)(79,0.4401)(80,0.4394)(81,0.4407)(82,0.4395)(83,0.4400)(84,0.4397)(85,0.4397)(86,0.4411)(87,0.4392)(88,0.4389)(89,0.4391)(90,0.4387)(91,0.4390)(92,0.4393)(93,0.4389)(94,0.4391)(95,0.4389)(96,0.4387)(97,0.4392)(98,0.4383)(99,0.4385)(100,0.4372)(101,0.4389)(102,0.4389)(103,0.4384)(104,0.4387)(105,0.4377)(106,0.4380)(107,0.4390)(108,0.4382)(109,0.4379)(110,0.4382)(111,0.4383)(112,0.4379)(113,0.4385)(114,0.4369)(115,0.4369)(116,0.4374)(117,0.4371)(118,0.4371)(119,0.4377)(120,0.4385)(121,0.4381)(122,0.4380)(123,0.4359)(124,0.4382)(125,0.4381)(126,0.4382)(127,0.4378)(128,0.4369)(129,0.4361)(130,0.4377)(131,0.4378)(132,0.4365)(133,0.4362)(134,0.4371)(135,0.4378)(136,0.4364)(137,0.4358)(138,0.4371)(139,0.4361)(140,0.4364)(141,0.4354)(142,0.4363)(143,0.4367)(144,0.4357)(145,0.4366)(146,0.4364)(147,0.4365)(148,0.4356)(149,0.4352)(150,0.4367)(151,0.4352)(152,0.4364)(153,0.4366)(154,0.4360)(155,0.4359)(156,0.4354)(157,0.4356)(158,0.4351)(159,0.4363)(160,0.4363)(161,0.4353)(162,0.4360)(163,0.4361)(164,0.4362)(165,0.4354)(166,0.4350)(167,0.4343)(168,0.4362)(169,0.4379)(170,0.4357)(171,0.4350)(172,0.4346)(173,0.4343)(174,0.4352)(175,0.4360)(176,0.4344)(177,0.4364)(178,0.4355)(179,0.4346)(180,0.4352)(181,0.4340)(182,0.4342)(183,0.4336)(184,0.4338)(185,0.4349)(186,0.4343)(187,0.4350)(188,0.4351)(189,0.4340)(190,0.4349)(191,0.4341)(192,0.4349)(193,0.4342)(194,0.4343)(195,0.4355)(196,0.4340)(197,0.4342)(198,0.4342)(199,0.4354)(200,0.4344)(201,0.4334)(202,0.4347)(203,0.4338)(204,0.4340)(205,0.4334)(206,0.4334)(207,0.4345)(208,0.4338)(209,0.4359)(210,0.4347)(211,0.4336)(212,0.4344)(213,0.4335)(214,0.4332)(215,0.4338)(216,0.4335)(217,0.4334)(218,0.4337)(219,0.4338)(220,0.4337)(221,0.4342)(222,0.4340)(223,0.4333)(224,0.4357)(225,0.4345)(226,0.4341)(227,0.4338)(228,0.4335)(229,0.4340)(230,0.4334)(231,0.4333)(232,0.4339)(233,0.4334)(234,0.4364)(235,0.4339)(236,0.4351)(237,0.4339)(238,0.4337)(239,0.4332)(240,0.4337)(241,0.4353)(242,0.4336)(243,0.4336)(244,0.4330)(245,0.4336)(246,0.4368)(247,0.4356)(248,0.4336)(249,0.4351)(250,0.4337)(251,0.4362)(252,0.4353)(253,0.4342)(254,0.4333)(255,0.4352)(256,0.4348)(257,0.4332)(258,0.4328)(259,0.4338)(260,0.4344)(261,0.4343)(262,0.4328)(263,0.4339)(264,0.4330)(265,0.4333)(266,0.4328)(267,0.4344)(268,0.4339)(269,0.4334)(270,0.4341)(271,0.4348)(272,0.4335)(273,0.4337)(274,0.4334)(275,0.4341)(276,0.4329)(277,0.4332)(278,0.4342)(279,0.4355)(280,0.4340)(281,0.4341)(282,0.4339)(283,0.4339)(284,0.4366)(285,0.4332)(286,0.4336)(287,0.4342)(288,0.4332)(289,0.4343)(290,0.4334)(291,0.4340)(292,0.4351)(293,0.4337)(294,0.4326)(295,0.4326)(296,0.4326)(297,0.4326)(298,0.4326)(299,0.4326)
    };   

\end{axis}
\end{tikzpicture}
\caption{The convergence behaviors of FM and GEM on RMSE}
\label{fig:comparison_rmse}
\end{figure}

\pgfplotsset{width=4.4cm, height=4.0cm, compat=1.5}
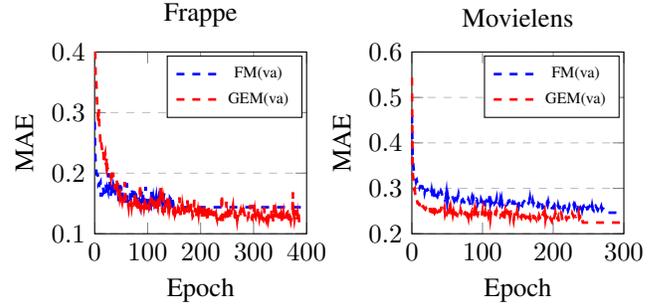
\begin{figure}[tb]
\begin{tikzpicture}
\begin{axis}[
    title={Frappe},
    xlabel={Epoch},
    ylabel={MAE},
    xmin=0, xmax=400,
    ymin=0.1, ymax=0.4,
    xtick={0,100,200,300,400},
    ytick={0.1,0.2,0.3,0.4},
    ymajorgrids=true,
    xmajorgrids=false,
    grid style=dashed,
    legend entries={FM(va), GEM(va)},
    legend style={
            font=\tiny,
            /tikz/every even column/.append style={column sep=0.5cm}
        }
]
\addplot[
    color=blue,
      line width=1pt,
    dashed = false,
    ]
    coordinates {
(0,0.2846)(1,0.2359)(2,0.2145)(3,0.2049)(4,0.1977)(5,0.1972)(6,0.1969)(7,0.1807)(8,0.1821)(9,0.1808)(10,0.1838)(11,0.1855)(12,0.1612)(13,0.1681)(14,0.1683)(15,0.1778)(16,0.1766)(17,0.1683)(18,0.1701)(19,0.1693)(20,0.1767)(21,0.1827)(22,0.1735)(23,0.1644)(24,0.1746)(25,0.1712)(26,0.2000)(27,0.1770)(28,0.1785)(29,0.1919)(30,0.1606)(31,0.1690)(32,0.1708)(33,0.1608)(34,0.1747)(35,0.1802)(36,0.1751)(37,0.1803)(38,0.1718)(39,0.1656)(40,0.1603)(41,0.1658)(42,0.1673)(43,0.1794)(44,0.1645)(45,0.1610)(46,0.1540)(47,0.1658)(48,0.1805)(49,0.1547)(50,0.1780)(51,0.1561)(52,0.1741)(53,0.1643)(54,0.1568)(55,0.1553)(56,0.1592)(57,0.1578)(58,0.1518)(59,0.1509)(60,0.1587)(61,0.1831)(62,0.1554)(63,0.1533)(64,0.1540)(65,0.1567)(66,0.1554)(67,0.1666)(68,0.1555)(69,0.1630)(70,0.1637)(71,0.1713)(72,0.1610)(73,0.1539)(74,0.1674)(75,0.1678)(76,0.1563)(77,0.1658)(78,0.1652)(79,0.1572)(80,0.1570)(81,0.1534)(82,0.1593)(83,0.1562)(84,0.1574)(85,0.1541)(86,0.1545)(87,0.1494)(88,0.1697)(89,0.1559)(90,0.1492)(91,0.1703)(92,0.1547)(93,0.1623)(94,0.1492)(95,0.1531)(96,0.1775)(97,0.1531)(98,0.1491)(99,0.1543)(100,0.1554)(101,0.1487)(102,0.1569)(103,0.1467)(104,0.1547)(105,0.1482)(106,0.1461)(107,0.1684)(108,0.1546)(109,0.1526)(110,0.1510)(111,0.1595)(112,0.1534)(113,0.1599)(114,0.1524)(115,0.1557)(116,0.1608)(117,0.1464)(118,0.1535)(119,0.1479)(120,0.1539)(121,0.1454)(122,0.1464)(123,0.1621)(124,0.1586)(125,0.1581)(126,0.1517)(127,0.1555)(128,0.1500)(129,0.1648)(130,0.1567)(131,0.1665)(132,0.1475)(133,0.1580)(134,0.1452)(135,0.1474)(136,0.1486)(137,0.1526)(138,0.1502)(139,0.1495)(140,0.1546)(141,0.1485)(142,0.1589)(143,0.1573)(144,0.1595)(145,0.1440)(146,0.1618)(147,0.1555)(148,0.1522)(149,0.1438)(150,0.1438)(151,0.1438)(152,0.1438)(153,0.1438)(154,0.1438)(155,0.1438)(156,0.1438)(157,0.1438)(158,0.1438)(159,0.1438)(160,0.1438)(161,0.1438)(162,0.1438)(163,0.1438)(164,0.1438)(165,0.1438)(166,0.1438)(167,0.1438)(168,0.1438)(169,0.1438)(170,0.1438)(171,0.1438)(172,0.1438)(173,0.1438)(174,0.1438)(175,0.1438)(176,0.1438)(177,0.1438)(178,0.1438)(179,0.1438)(180,0.1438)(181,0.1438)(182,0.1438)(183,0.1438)(184,0.1438)(185,0.1438)(186,0.1438)(187,0.1438)(188,0.1438)(189,0.1438)(190,0.1438)(191,0.1438)(192,0.1438)(193,0.1438)(194,0.1438)(195,0.1438)(196,0.1438)(197,0.1438)(198,0.1438)(199,0.1438)(200,0.1438)(201,0.1438)(202,0.1438)(203,0.1438)(204,0.1438)(205,0.1438)(206,0.1438)(207,0.1438)(208,0.1438)(209,0.1438)(210,0.1438)(211,0.1438)(212,0.1438)(213,0.1438)(214,0.1438)(215,0.1438)(216,0.1438)(217,0.1438)(218,0.1438)(219,0.1438)(220,0.1438)(221,0.1438)(222,0.1438)(223,0.1438)(224,0.1438)(225,0.1438)(226,0.1438)(227,0.1438)(228,0.1438)(229,0.1438)(230,0.1438)(231,0.1438)(232,0.1438)(233,0.1438)(234,0.1438)(235,0.1438)(236,0.1438)(237,0.1438)(238,0.1438)(239,0.1438)(240,0.1438)(241,0.1438)(242,0.1438)(243,0.1438)(244,0.1438)(245,0.1438)(246,0.1438)(247,0.1438)(248,0.1438)(249,0.1438)(250,0.1438)(251,0.1438)(252,0.1438)(253,0.1438)(254,0.1438)(255,0.1438)(256,0.1438)(257,0.1438)(258,0.1438)(259,0.1438)(260,0.1438)(261,0.1438)(262,0.1438)(263,0.1438)(264,0.1438)(265,0.1438)(266,0.1438)(267,0.1438)(268,0.1438)(269,0.1438)(270,0.1438)(271,0.1438)(272,0.1438)(273,0.1438)(274,0.1438)(275,0.1438)(276,0.1438)(277,0.1438)(278,0.1438)(279,0.1438)(280,0.1438)(281,0.1438)(282,0.1438)(283,0.1438)(284,0.1438)(285,0.1438)(286,0.1438)(287,0.1438)(288,0.1438)(289,0.1438)(290,0.1438)(291,0.1438)(292,0.1438)(293,0.1438)(294,0.1438)(295,0.1438)(296,0.1438)(297,0.1438)(298,0.1438)(299,0.1438)(300,0.1438)(301,0.1438)(302,0.1438)(303,0.1438)(304,0.1438)(305,0.1438)(306,0.1438)(307,0.1438)(308,0.1438)(309,0.1438)(310,0.1438)(311,0.1438)(312,0.1438)(313,0.1438)(314,0.1438)(315,0.1438)(316,0.1438)(317,0.1438)(318,0.1438)(319,0.1438)(320,0.1438)(321,0.1438)(322,0.1438)(323,0.1438)(324,0.1438)(325,0.1438)(326,0.1438)(327,0.1438)(328,0.1438)(329,0.1438)(330,0.1438)(331,0.1438)(332,0.1438)(333,0.1438)(334,0.1438)(335,0.1438)(336,0.1438)(337,0.1438)(338,0.1438)(339,0.1438)(340,0.1438)(341,0.1438)(342,0.1438)(343,0.1438)(344,0.1438)(345,0.1438)(346,0.1438)(347,0.1438)(348,0.1438)(349,0.1438)(350,0.1438)(351,0.1438)(352,0.1438)(353,0.1438)(354,0.1438)(355,0.1438)(356,0.1438)(357,0.1438)(358,0.1438)(359,0.1438)(360,0.1438)(361,0.1438)(362,0.1438)(363,0.1438)(364,0.1438)(365,0.1438)(366,0.1438)(367,0.1438)(368,0.1438)(369,0.1438)(370,0.1438)(371,0.1438)(372,0.1438)(373,0.1438)(374,0.1438)(375,0.1438)(376,0.1438)(377,0.1438)(378,0.1438)(379,0.1438)(380,0.1438)(381,0.1438)(382,0.1438)(383,0.1438)(384,0.1438)(385,0.1438)(386,0.1438)(387,0.1438)(388,0.1438)(389,0.1438)(390,0.1438)(391,0.1438)(392,0.1438)(393,0.1438)(394,0.1438)(395,0.1438)(396,0.1438)(397,0.1438)(398,0.1438)(399,0.1438)
    }; 

\addplot[
    color=red,
      line width=1pt,
    dashed = false,
    ]
    coordinates {
(0,0.5059)(1,0.4055)(2,0.3509)(3,0.3524)(4,0.3337)(5,0.3077)(6,0.3010)(7,0.2836)(8,0.3048)(9,0.2727)(10,0.2705)(11,0.2466)(12,0.2530)(13,0.2467)(14,0.2395)(15,0.2347)(16,0.2399)(17,0.2283)(18,0.2152)(19,0.2282)(20,0.2322)(21,0.2205)(22,0.2141)(23,0.2269)(24,0.2069)(25,0.2095)(26,0.2105)(27,0.2055)(28,0.2006)(29,0.2009)(30,0.2000)(31,0.1926)(32,0.1988)(33,0.1893)(34,0.1990)(35,0.1990)(36,0.1855)(37,0.1864)(38,0.1751)(39,0.1746)(40,0.1784)(41,0.1899)(42,0.1859)(43,0.1795)(44,0.1850)(45,0.1621)(46,0.1693)(47,0.1611)(48,0.1679)(49,0.1745)(50,0.1630)(51,0.1556)(52,0.1513)(53,0.1565)(54,0.1584)(55,0.1507)(56,0.1562)(57,0.1852)(58,0.1673)(59,0.1555)(60,0.1468)(61,0.1498)(62,0.1510)(63,0.1506)(64,0.1446)(65,0.1566)(66,0.1563)(67,0.1595)(68,0.1557)(69,0.1527)(70,0.1447)(71,0.1517)(72,0.1550)(73,0.1526)(74,0.1501)(75,0.1524)(76,0.1697)(77,0.1554)(78,0.1411)(79,0.1614)(80,0.1620)(81,0.1489)(82,0.1383)(83,0.1515)(84,0.1498)(85,0.1462)(86,0.1818)(87,0.1480)(88,0.1416)(89,0.1433)(90,0.1441)(91,0.1520)(92,0.1490)(93,0.1429)(94,0.1504)(95,0.1527)(96,0.1491)(97,0.1556)(98,0.1400)(99,0.1482)(100,0.1496)(101,0.1441)(102,0.1415)(103,0.1457)(104,0.1421)(105,0.1383)(106,0.1543)(107,0.1526)(108,0.1592)(109,0.1604)(110,0.1454)(111,0.1597)(112,0.1535)(113,0.1552)(114,0.1530)(115,0.1351)(116,0.1501)(117,0.1417)(118,0.1491)(119,0.1453)(120,0.1644)(121,0.1395)(122,0.1391)(123,0.1549)(124,0.1509)(125,0.1789)(126,0.1460)(127,0.1465)(128,0.1370)(129,0.1355)(130,0.1585)(131,0.1534)(132,0.1356)(133,0.1458)(134,0.1441)(135,0.1439)(136,0.1392)(137,0.1498)(138,0.1647)(139,0.1459)(140,0.1501)(141,0.1402)(142,0.1465)(143,0.1409)(144,0.1382)(145,0.1357)(146,0.1398)(147,0.1399)(148,0.1365)(149,0.1391)(150,0.1350)(151,0.1492)(152,0.1407)(153,0.1348)(154,0.1375)(155,0.1410)(156,0.1413)(157,0.1332)(158,0.1430)(159,0.1414)(160,0.1599)(161,0.1406)(162,0.1311)(163,0.1501)(164,0.1491)(165,0.1306)(166,0.1465)(167,0.1408)(168,0.1420)(169,0.1318)(170,0.1434)(171,0.1326)(172,0.1393)(173,0.1407)(174,0.1351)(175,0.1550)(176,0.1456)(177,0.1529)(178,0.1431)(179,0.1384)(180,0.1399)(181,0.1308)(182,0.1509)(183,0.1456)(184,0.1380)(185,0.1284)(186,0.1303)(187,0.1329)(188,0.1562)(189,0.1482)(190,0.1358)(191,0.1440)(192,0.1296)(193,0.1400)(194,0.1333)(195,0.1391)(196,0.1327)(197,0.1443)(198,0.1489)(199,0.1419)(200,0.1304)(201,0.1319)(202,0.1258)(203,0.1404)(204,0.1469)(205,0.1286)(206,0.1441)(207,0.1258)(208,0.1289)(209,0.1265)(210,0.1303)(211,0.1415)(212,0.1274)(213,0.1276)(214,0.1345)(215,0.1473)(216,0.1253)(217,0.1251)(218,0.1362)(219,0.1352)(220,0.1365)(221,0.1299)(222,0.1267)(223,0.1307)(224,0.1269)(225,0.1371)(226,0.1254)(227,0.1276)(228,0.1290)(229,0.1326)(230,0.1265)(231,0.1232)(232,0.1323)(233,0.1297)(234,0.1249)(235,0.1253)(236,0.1240)(237,0.1235)(238,0.1270)(239,0.1352)(240,0.1651)(241,0.1322)(242,0.1212)(243,0.1335)(244,0.1334)(245,0.1307)(246,0.1289)(247,0.1309)(248,0.1529)(249,0.1426)(250,0.1266)(251,0.1337)(252,0.1305)(253,0.1343)(254,0.1420)(255,0.1306)(256,0.1245)(257,0.1394)(258,0.1295)(259,0.1347)(260,0.1416)(261,0.1461)(262,0.1344)(263,0.1261)(264,0.1270)(265,0.1440)(266,0.1247)(267,0.1324)(268,0.1310)(269,0.1223)(270,0.1291)(271,0.1294)(272,0.1331)(273,0.1272)(274,0.1350)(275,0.1381)(276,0.1276)(277,0.1266)(278,0.1281)(279,0.1432)(280,0.1307)(281,0.1385)(282,0.1351)(283,0.1308)(284,0.1299)(285,0.1245)(286,0.1240)(287,0.1286)(288,0.1297)(289,0.1258)(290,0.1282)(291,0.1373)(292,0.1354)(293,0.1328)(294,0.1473)(295,0.1295)(296,0.1265)(297,0.1283)(298,0.1260)(299,0.1362)(300,0.1480)(301,0.1306)(302,0.1374)(303,0.1282)(304,0.1261)(305,0.1299)(306,0.1240)(307,0.1242)(308,0.1452)(309,0.1396)(310,0.1340)(311,0.1256)(312,0.1299)(313,0.1278)(314,0.1242)(315,0.1296)(316,0.1244)(317,0.1244)(318,0.1229)(319,0.1302)(320,0.1437)(321,0.1214)(322,0.1222)(323,0.1241)(324,0.1212)(325,0.1280)(326,0.1289)(327,0.1369)(328,0.1250)(329,0.1301)(330,0.1337)(331,0.1295)(332,0.1248)(333,0.1262)(334,0.1419)(335,0.1296)(336,0.1296)(337,0.1209)(338,0.1376)(339,0.1505)(340,0.1240)(341,0.1282)(342,0.1368)(343,0.1314)(344,0.1295)(345,0.1220)(346,0.1263)(347,0.1238)(348,0.1323)(349,0.1330)(350,0.1308)(351,0.1204)(352,0.1295)(353,0.1232)(354,0.1202)(355,0.1299)(356,0.1348)(357,0.1236)(358,0.1247)(359,0.1213)(360,0.1240)(361,0.1221)(362,0.1330)(363,0.1346)(364,0.1355)(365,0.1281)(366,0.1216)(367,0.1326)(368,0.1198)(369,0.1247)(370,0.1207)(371,0.1271)(372,0.1269)(373,0.1333)(374,0.1682)(375,0.1246)(376,0.1320)(377,0.1236)(378,0.1298)(379,0.1269)(380,0.1387)(381,0.1209)(382,0.1227)(383,0.1308)(384,0.1215)(385,0.1319)(386,0.1186)(387,0.1186)(388,0.1186)(389,0.1186)(390,0.1186)(391,0.1186)(392,0.1186)(393,0.1186)(394,0.1186)(395,0.1186)(396,0.1186)(397,0.1186)(398,0.1186)(399,0.1186)

    };  
\end{axis}

\hskip 120pt
\begin{axis}[
    title={Movielens},
    xlabel={Epoch},
    ylabel={MAE},
    xmin=0, xmax=300,
    ymin=0.2, ymax=0.6,
    xtick={0,100,200,300},
    ytick={0.2,0.3,0.4,0.5,0.6},
    ymajorgrids=true,
    xmajorgrids=false,
    grid style=dashed,
    legend entries={ FM(va),  GEM(va)},
    legend style={
            font=\tiny,
            /tikz/every even column/.append style={column sep=0.5cm}
        }
]

\addplot[
    color=blue,
      line width=1pt,
    dashed = true,
    ]
    coordinates {
 (0,0.4607)(1,0.3719)(2,0.3494)(3,0.3358)(4,0.3282)(5,0.3215)(6,0.3130)(7,0.3168)(8,0.3091)(9,0.3125)(10,0.3106)(11,0.3013)(12,0.3046)(13,0.3038)(14,0.3046)(15,0.3041)(16,0.2952)(17,0.3003)(18,0.2898)(19,0.2967)(20,0.2961)(21,0.2958)(22,0.2975)(23,0.2954)(24,0.2909)(25,0.2922)(26,0.2886)(27,0.2855)(28,0.2959)(29,0.2856)(30,0.2860)(31,0.3008)(32,0.2780)(33,0.2900)(34,0.2871)(35,0.2833)(36,0.2812)(37,0.2872)(38,0.2821)(39,0.2776)(40,0.2901)(41,0.2947)(42,0.2912)(43,0.2839)(44,0.2819)(45,0.2815)(46,0.2849)(47,0.2754)(48,0.2781)(49,0.2901)(50,0.2699)(51,0.2811)(52,0.2710)(53,0.2751)(54,0.2742)(55,0.2757)(56,0.2789)(57,0.2798)(58,0.2800)(59,0.2755)(60,0.2714)(61,0.2749)(62,0.2709)(63,0.2838)(64,0.2765)(65,0.2803)(66,0.2878)(67,0.2799)(68,0.2768)(69,0.2759)(70,0.2709)(71,0.2744)(72,0.2756)(73,0.2701)(74,0.2694)(75,0.2765)(76,0.2722)(77,0.2709)(78,0.2723)(79,0.2709)(80,0.2767)(81,0.2735)(82,0.2683)(83,0.2618)(84,0.2726)(85,0.2797)(86,0.2718)(87,0.2775)(88,0.2687)(89,0.2721)(90,0.2736)(91,0.2771)(92,0.2761)(93,0.2650)(94,0.2697)(95,0.2713)(96,0.2638)(97,0.2625)(98,0.2757)(99,0.2716)(100,0.2698)(101,0.2578)(102,0.2705)(103,0.2708)(104,0.2573)(105,0.2648)(106,0.2725)(107,0.2708)(108,0.2712)(109,0.2736)(110,0.2837)(111,0.2656)(112,0.2725)(113,0.2677)(114,0.2700)(115,0.2791)(116,0.2779)(117,0.2666)(118,0.2689)(119,0.2707)(120,0.2778)(121,0.2626)(122,0.2662)(123,0.2663)(124,0.2639)(125,0.2740)(126,0.2637)(127,0.2742)(128,0.2619)(129,0.2647)(130,0.2706)(131,0.2706)(132,0.2695)(133,0.2703)(134,0.2637)(135,0.2655)(136,0.2683)(137,0.2729)(138,0.2733)(139,0.2672)(140,0.2636)(141,0.2578)(142,0.2649)(143,0.2701)(144,0.2681)(145,0.2695)(146,0.2654)(147,0.2599)(148,0.2704)(149,0.2666)(150,0.2653)(151,0.2605)(152,0.2731)(153,0.2633)(154,0.2654)(155,0.2624)(156,0.2613)(157,0.2680)(158,0.2692)(159,0.2716)(160,0.2706)(161,0.2673)(162,0.2646)(163,0.2618)(164,0.2691)(165,0.2669)(166,0.2692)(167,0.2782)(168,0.2641)(169,0.2644)(170,0.2633)(171,0.2731)(172,0.2664)(173,0.2713)(174,0.2632)(175,0.2647)(176,0.2655)(177,0.2618)(178,0.2598)(179,0.2659)(180,0.2632)(181,0.2716)(182,0.2745)(183,0.2673)(184,0.2683)(185,0.2567)(186,0.2576)(187,0.2585)(188,0.2678)(189,0.2642)(190,0.2615)(191,0.2667)(192,0.2625)(193,0.2705)(194,0.2623)(195,0.2663)(196,0.2646)(197,0.2624)(198,0.2566)(199,0.2568)(200,0.2594)(201,0.2571)(202,0.2633)(203,0.2604)(204,0.2613)(205,0.2585)(206,0.2537)(207,0.2631)(208,0.2568)(209,0.2563)(210,0.2630)(211,0.2566)(212,0.2609)(213,0.2603)(214,0.2545)(215,0.2606)(216,0.2613)(217,0.2517)(218,0.2573)(219,0.2551)(220,0.2585)(221,0.2513)(222,0.2558)(223,0.2621)(224,0.2580)(225,0.2596)(226,0.2630)(227,0.2599)(228,0.2601)(229,0.2642)(230,0.2578)(231,0.2629)(232,0.2579)(233,0.2651)(234,0.2544)(235,0.2578)(236,0.2568)(237,0.2552)(238,0.2574)(239,0.2515)(240,0.2618)(241,0.2520)(242,0.2593)(243,0.2610)(244,0.2555)(245,0.2576)(246,0.2652)(247,0.2593)(248,0.2543)(249,0.2547)(250,0.2588)(251,0.2603)(252,0.2526)(253,0.2560)(254,0.2531)(255,0.2622)(256,0.2531)(257,0.2513)(258,0.2645)(259,0.2595)(260,0.2613)(261,0.2582)(262,0.2548)(263,0.2518)(264,0.2557)(265,0.2594)(266,0.2642)(267,0.2553)(268,0.2599)(269,0.2528)(270,0.2557)(271,0.2552)(272,0.2464)(273,0.2464)(274,0.2464)(275,0.2464)(276,0.2464)(277,0.2464)(278,0.2464)(279,0.2464)(280,0.2464)(281,0.2464)(282,0.2464)(283,0.2464)(284,0.2464)(285,0.2464)(286,0.2464)(287,0.2464)(288,0.2464)(289,0.2464)(290,0.2464)(291,0.2464)(292,0.2464)(293,0.2464)(294,0.2464)(295,0.2464)(296,0.2464)(297,0.2464)(298,0.2464)(299,0.2464)
    }; 

\addplot[
    color=red,
      line width=1pt,
    dashed = true,
    ]
    coordinates {
  (0,0.5435)(1,0.3368)(2,0.3227)(3,0.2990)(4,0.3003)(5,0.2802)(6,0.2829)(7,0.2749)(8,0.2676)(9,0.2679)(10,0.2651)(11,0.2649)(12,0.2549)(13,0.2567)(14,0.2640)(15,0.2604)(16,0.2609)(17,0.2570)(18,0.2553)(19,0.2535)(20,0.2562)(21,0.2525)(22,0.2590)(23,0.2565)(24,0.2590)(25,0.2579)(26,0.2492)(27,0.2489)(28,0.2592)(29,0.2445)(30,0.2490)(31,0.2471)(32,0.2645)(33,0.2484)(34,0.2512)(35,0.2489)(36,0.2471)(37,0.2480)(38,0.2375)(39,0.2508)(40,0.2501)(41,0.2520)(42,0.2606)(43,0.2507)(44,0.2518)(45,0.2522)(46,0.2575)(47,0.2426)(48,0.2484)(49,0.2524)(50,0.2449)(51,0.2484)(52,0.2474)(53,0.2544)(54,0.2413)(55,0.2549)(56,0.2425)(57,0.2514)(58,0.2412)(59,0.2493)(60,0.2409)(61,0.2531)(62,0.2417)(63,0.2411)(64,0.2398)(65,0.2464)(66,0.2437)(67,0.2451)(68,0.2386)(69,0.2528)(70,0.2475)(71,0.2531)(72,0.2467)(73,0.2500)(74,0.2451)(75,0.2464)(76,0.2434)(77,0.2412)(78,0.2538)(79,0.2419)(80,0.2348)(81,0.2490)(82,0.2411)(83,0.2433)(84,0.2384)(85,0.2500)(86,0.2535)(87,0.2512)(88,0.2377)(89,0.2516)(90,0.2396)(91,0.2497)(92,0.2431)(93,0.2404)(94,0.2445)(95,0.2463)(96,0.2450)(97,0.2383)(98,0.2415)(99,0.2381)(100,0.2500)(101,0.2328)(102,0.2373)(103,0.2364)(104,0.2471)(105,0.2355)(106,0.2454)(107,0.2480)(108,0.2406)(109,0.2501)(110,0.2495)(111,0.2333)(112,0.2468)(113,0.2448)(114,0.2451)(115,0.2395)(116,0.2356)(117,0.2350)(118,0.2460)(119,0.2436)(120,0.2506)(121,0.2437)(122,0.2457)(123,0.2456)(124,0.2442)(125,0.2385)(126,0.2417)(127,0.2405)(128,0.2397)(129,0.2502)(130,0.2335)(131,0.2358)(132,0.2476)(133,0.2358)(134,0.2409)(135,0.2475)(136,0.2437)(137,0.2432)(138,0.2541)(139,0.2354)(140,0.2432)(141,0.2479)(142,0.2467)(143,0.2389)(144,0.2298)(145,0.2414)(146,0.2456)(147,0.2448)(148,0.2346)(149,0.2491)(150,0.2421)(151,0.2494)(152,0.2387)(153,0.2378)(154,0.2378)(155,0.2349)(156,0.2410)(157,0.2400)(158,0.2515)(159,0.2401)(160,0.2360)(161,0.2344)(162,0.2374)(163,0.2375)(164,0.2477)(165,0.2349)(166,0.2336)(167,0.2383)(168,0.2323)(169,0.2368)(170,0.2353)(171,0.2445)(172,0.2355)(173,0.2343)(174,0.2368)(175,0.2351)(176,0.2318)(177,0.2415)(178,0.2358)(179,0.2334)(180,0.2406)(181,0.2364)(182,0.2296)(183,0.2356)(184,0.2334)(185,0.2416)(186,0.2456)(187,0.2424)(188,0.2489)(189,0.2306)(190,0.2393)(191,0.2356)(192,0.2371)(193,0.2434)(194,0.2368)(195,0.2314)(196,0.2343)(197,0.2304)(198,0.2275)(199,0.2408)(200,0.2311)(201,0.2337)(202,0.2386)(203,0.2457)(204,0.2331)(205,0.2279)(206,0.2330)(207,0.2373)(208,0.2266)(209,0.2362)(210,0.2401)(211,0.2451)(212,0.2411)(213,0.2293)(214,0.2339)(215,0.2296)(216,0.2344)(217,0.2336)(218,0.2360)(219,0.2317)(220,0.2345)(221,0.2358)(222,0.2365)(223,0.2325)(224,0.2328)(225,0.2380)(226,0.2357)(227,0.2446)(228,0.2408)(229,0.2370)(230,0.2391)(231,0.2394)(232,0.2379)(233,0.2408)(234,0.2413)(235,0.2368)(236,0.2454)(237,0.2337)(238,0.2363)(239,0.2318)(240,0.2383)(241,0.2317)(242,0.2246)(243,0.2246)(244,0.2246)(245,0.2246)(246,0.2246)(247,0.2246)(248,0.2246)(249,0.2246)(250,0.2246)(251,0.2246)(252,0.2246)(253,0.2246)(254,0.2246)(255,0.2246)(256,0.2246)(257,0.2246)(258,0.2246)(259,0.2246)(260,0.2246)(261,0.2246)(262,0.2246)(263,0.2246)(264,0.2246)(265,0.2246)(266,0.2246)(267,0.2246)(268,0.2246)(269,0.2246)(270,0.2246)(271,0.2246)(272,0.2246)(273,0.2246)(274,0.2246)(275,0.2246)(276,0.2246)(277,0.2246)(278,0.2246)(279,0.2246)(280,0.2246)(281,0.2246)(282,0.2246)(283,0.2246)(284,0.2246)(285,0.2246)(286,0.2246)(287,0.2246)(288,0.2246)(289,0.2246)(290,0.2246)(291,0.2246)(292,0.2246)(293,0.2246)(294,0.2246)(295,0.2246)(296,0.2246)(297,0.2246)(298,0.2246)(299,0.2246)
    };   

\end{axis}
\end{tikzpicture}
\caption{The convergence behaviors of FM and GEM on MAE}
\label{fig:comparison_mae}
\end{figure}

\subsection{(RQ2) Graph Constructing and Neighbor Sampling Ratio}
\subsubsection{The Impact of Graph Constructing}
The way to construct graphs based on the feature vector plays an important role on GEM because it affects the information propagation and the learning of high-order embeddings. We propose two kinds of method to construct graphs in section 3.2 (see Figure \ref{fig:gem-graph} for details). Here we conduct experiments to see the performance of the two methods. The results in terms of RMSE are described in Figure \ref{fig:nodetype}.
First, by comparing FM (the black line) and GEM(type:b) (the blue line), we can see that building higher-order embedding just from user-item edges can even improve the model performance. Based on such observation, we add some features to the graph. More precisely, on Frappe dataset we add the features of cities and countries. On Movielens dataset we add the features of tags. Then we can get GEM(type:a) (the red line). It's obvious that GEM(type:a) achieves more stable convergence and or better final performance. We can assume that the proposed GEM could do better when we add more suitable features into the graph. However, if we add  bad features ($isweekend$ in this example) 
to the graph, the performance would downgrade a lot as shown in the green line on the Frappe dataset.
The reason is that the feature $isweekend$ only has two nodes so the node of this feature actually connects with almost half of the other nodes. It would reduce the discrimination of features embeddings after graph convolution.

\pgfplotsset{width=4.4cm, height=4.0cm, compat=1.5}
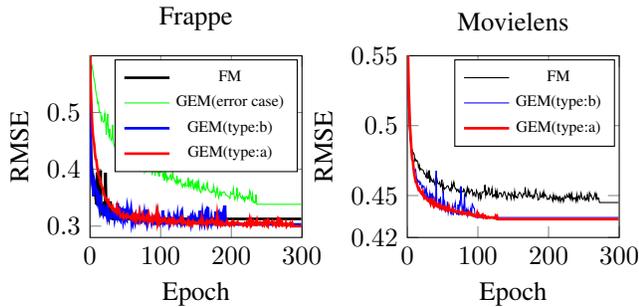
\begin{figure}[tb]

\begin{tikzpicture}
\begin{axis}[
    title={Frappe},
    xlabel={Epoch},
    ylabel={RMSE},
    xmin=0, xmax=300,
    ymin=0.28, ymax=0.6,
    xtick={0,100,200,300},
    ytick={0.3,0.4,0.5},
    xmajorgrids=false,
    grid style=dashed,
    legend entries={FM, GEM(error case), GEM(type:b), GEM(type:a)},
    legend style={
            font=\tiny,
            /tikz/every even column/.append style={column sep=0.5cm}}
]

\addplot[
    color=black,
    line width=1pt,
    ]
    coordinates {
       (0,0.6908)(1,0.4661)(2,0.4117)(3,0.3963)(4,0.3775)(5,0.3803)(6,0.3589)(7,0.3712)(8,0.3502)(9,0.3386)(10,0.3945)(11,0.3360)(12,0.3693)(13,0.3391)(14,0.3374)(15,0.3981)(16,0.3234)(17,0.3257)(18,0.3383)(19,0.3203)(20,0.3427)(21,0.3177)(22,0.3941)(23,0.3347)(24,0.3175)(25,0.3166)(26,0.3265)(27,0.3182)(28,0.3303)(29,0.3436)(30,0.3279)(31,0.3155)(32,0.3127)(33,0.3127)(34,0.3127)(35,0.3127)(36,0.3127)(37,0.3127)(38,0.3127)(39,0.3127)(40,0.3127)(41,0.3127)(42,0.3127)(43,0.3127)(44,0.3127)(45,0.3127)(46,0.3127)(47,0.3127)(48,0.3127)(49,0.3127)(50,0.3127)(51,0.3127)(52,0.3127)(53,0.3127)(54,0.3127)(55,0.3127)(56,0.3127)(57,0.3127)(58,0.3127)(59,0.3127)(60,0.3127)(61,0.3127)(62,0.3127)(63,0.3127)(64,0.3127)(65,0.3127)(66,0.3127)(67,0.3127)(68,0.3127)(69,0.3127)(70,0.3127)(71,0.3127)(72,0.3127)(73,0.3127)(74,0.3127)(75,0.3127)(76,0.3127)(77,0.3127)(78,0.3127)(79,0.3127)(80,0.3127)(81,0.3127)(82,0.3127)(83,0.3127)(84,0.3127)(85,0.3127)(86,0.3127)(87,0.3127)(88,0.3127)(89,0.3127)(90,0.3127)(91,0.3127)(92,0.3127)(93,0.3127)(94,0.3127)(95,0.3127)(96,0.3127)(97,0.3127)(98,0.3127)(99,0.3127)(100,0.3127)(101,0.3127)(102,0.3127)(103,0.3127)(104,0.3127)(105,0.3127)(106,0.3127)(107,0.3127)(108,0.3127)(109,0.3127)(110,0.3127)(111,0.3127)(112,0.3127)(113,0.3127)(114,0.3127)(115,0.3127)(116,0.3127)(117,0.3127)(118,0.3127)(119,0.3127)(120,0.3127)(121,0.3127)(122,0.3127)(123,0.3127)(124,0.3127)(125,0.3127)(126,0.3127)(127,0.3127)(128,0.3127)(129,0.3127)(130,0.3127)(131,0.3127)(132,0.3127)(133,0.3127)(134,0.3127)(135,0.3127)(136,0.3127)(137,0.3127)(138,0.3127)(139,0.3127)(140,0.3127)(141,0.3127)(142,0.3127)(143,0.3127)(144,0.3127)(145,0.3127)(146,0.3127)(147,0.3127)(148,0.3127)(149,0.3127)(150,0.3127)(151,0.3127)(152,0.3127)(153,0.3127)(154,0.3127)(155,0.3127)(156,0.3127)(157,0.3127)(158,0.3127)(159,0.3127)(160,0.3127)(161,0.3127)(162,0.3127)(163,0.3127)(164,0.3127)(165,0.3127)(166,0.3127)(167,0.3127)(168,0.3127)(169,0.3127)(170,0.3127)(171,0.3127)(172,0.3127)(173,0.3127)(174,0.3127)(175,0.3127)(176,0.3127)(177,0.3127)(178,0.3127)(179,0.3127)(180,0.3127)(181,0.3127)(182,0.3127)(183,0.3127)(184,0.3127)(185,0.3127)(186,0.3127)(187,0.3127)(188,0.3127)(189,0.3127)(190,0.3127)(191,0.3127)(192,0.3127)(193,0.3127)(194,0.3127)(195,0.3127)(196,0.3127)(197,0.3127)(198,0.3127)(199,0.3127)(200,0.3127)(201,0.3127)(202,0.3127)(203,0.3127)(204,0.3127)(205,0.3127)(206,0.3127)(207,0.3127)(208,0.3127)(209,0.3127)(210,0.3127)(211,0.3127)(212,0.3127)(213,0.3127)(214,0.3127)(215,0.3127)(216,0.3127)(217,0.3127)(218,0.3127)(219,0.3127)(220,0.3127)(221,0.3127)(222,0.3127)(223,0.3127)(224,0.3127)(225,0.3127)(226,0.3127)(227,0.3127)(228,0.3127)(229,0.3127)(230,0.3127)(231,0.3127)(232,0.3127)(233,0.3127)(234,0.3127)(235,0.3127)(236,0.3127)(237,0.3127)(238,0.3127)(239,0.3127)(240,0.3127)(241,0.3127)(242,0.3127)(243,0.3127)(244,0.3127)(245,0.3127)(246,0.3127)(247,0.3127)(248,0.3127)(249,0.3127)(250,0.3127)(251,0.3127)(252,0.3127)(253,0.3127)(254,0.3127)(255,0.3127)(256,0.3127)(257,0.3127)(258,0.3127)(259,0.3127)(260,0.3127)(261,0.3127)(262,0.3127)(263,0.3127)(264,0.3127)(265,0.3127)(266,0.3127)(267,0.3127)(268,0.3127)(269,0.3127)(270,0.3127)(271,0.3127)(272,0.3127)(273,0.3127)(274,0.3127)(275,0.3127)(276,0.3127)(277,0.3127)(278,0.3127)(279,0.3127)(280,0.3127)(281,0.3127)(282,0.3127)(283,0.3127)(284,0.3127)(285,0.3127)(286,0.3127)(287,0.3127)(288,0.3127)(289,0.3127)(290,0.3127)(291,0.3127)(292,0.3127)(293,0.3127)(294,0.3127)(295,0.3127)(296,0.3127)(297,0.3127)(298,0.3127)(299,0.3127)
    }; 

\addplot[
    color=green,
    ]
    coordinates {
    (0,0.6294)(1,0.5974)(2,0.5868)(3,0.5805)(4,0.5736)(5,0.5642)(6,0.5576)(7,0.5516)(8,0.5391)(9,0.5328)(10,0.5244)(11,0.5161)(12,0.5183)(13,0.5018)(14,0.5017)(15,0.4980)(16,0.4866)(17,0.4921)(18,0.4850)(19,0.4831)(20,0.4846)(21,0.4984)(22,0.4874)(23,0.4611)(24,0.4807)(25,0.4974)(26,0.4668)(27,0.4580)(28,0.4714)(29,0.4570)(30,0.4508)(31,0.4534)(32,0.4782)(33,0.4484)(34,0.4429)(35,0.4394)(36,0.4437)(37,0.4355)(38,0.4470)(39,0.4650)(40,0.4419)(41,0.4318)(42,0.4335)(43,0.4344)(44,0.4379)(45,0.4326)(46,0.4274)(47,0.4527)(48,0.4267)(49,0.4294)(50,0.4340)(51,0.4233)(52,0.4187)(53,0.4492)(54,0.4221)(55,0.4328)(56,0.4321)(57,0.4064)(58,0.4255)(59,0.4199)(60,0.4113)(61,0.4596)(62,0.4260)(63,0.4108)(64,0.4240)(65,0.4244)(66,0.4119)(67,0.4227)(68,0.4119)(69,0.4085)(70,0.4145)(71,0.4021)(72,0.4107)(73,0.4047)(74,0.4092)(75,0.4043)(76,0.4045)(77,0.3962)(78,0.4067)(79,0.3962)(80,0.3989)(81,0.4002)(82,0.3974)(83,0.4033)(84,0.3971)(85,0.3973)(86,0.3975)(87,0.4005)(88,0.3920)(89,0.3960)(90,0.3974)(91,0.4257)(92,0.3890)(93,0.4039)(94,0.3988)(95,0.4100)(96,0.3943)(97,0.3997)(98,0.3871)(99,0.4002)(100,0.3950)(101,0.3971)(102,0.4056)(103,0.3881)(104,0.3913)(105,0.3802)(106,0.4117)(107,0.3853)(108,0.3906)(109,0.3800)(110,0.3858)(111,0.3824)(112,0.3828)(113,0.3865)(114,0.3900)(115,0.3814)(116,0.3815)(117,0.3978)(118,0.3846)(119,0.3840)(120,0.3786)(121,0.3795)(122,0.3755)(123,0.3890)(124,0.3758)(125,0.3810)(126,0.3770)(127,0.3789)(128,0.3822)(129,0.3756)(130,0.3716)(131,0.3688)(132,0.3744)(133,0.3737)(134,0.3698)(135,0.3763)(136,0.3712)(137,0.3702)(138,0.3708)(139,0.3852)(140,0.3693)(141,0.3713)(142,0.3780)(143,0.3696)(144,0.3783)(145,0.3807)(146,0.3674)(147,0.3651)(148,0.3744)(149,0.3683)(150,0.3746)(151,0.3666)(152,0.3799)(153,0.3676)(154,0.3779)(155,0.3629)(156,0.3609)(157,0.3590)(158,0.3674)(159,0.3702)(160,0.3644)(161,0.3732)(162,0.3667)(163,0.3628)(164,0.3638)(165,0.3697)(166,0.3761)(167,0.3649)(168,0.3641)(169,0.3604)(170,0.3604)(171,0.3596)(172,0.3586)(173,0.3624)(174,0.3603)(175,0.3707)(176,0.3605)(177,0.3611)(178,0.3530)(179,0.3584)(180,0.3670)(181,0.3580)(182,0.3498)(183,0.3543)(184,0.3608)(185,0.3544)(186,0.3534)(187,0.3597)(188,0.3588)(189,0.3600)(190,0.3603)(191,0.3589)(192,0.3550)(193,0.3544)(194,0.3537)(195,0.3530)(196,0.3519)(197,0.3516)(198,0.3490)(199,0.3594)(200,0.3565)(201,0.3560)(202,0.3564)(203,0.3662)(204,0.3530)(205,0.3472)(206,0.3512)(207,0.3518)(208,0.3673)(209,0.3486)(210,0.3540)(211,0.3550)(212,0.3582)(213,0.3511)(214,0.3509)(215,0.3487)(216,0.3528)(217,0.3471)(218,0.3506)(219,0.3438)(220,0.3490)(221,0.3520)(222,0.3517)(223,0.3482)(224,0.3522)(225,0.3481)(226,0.3625)(227,0.3584)(228,0.3506)(229,0.3455)(230,0.3561)(231,0.3422)(232,0.3437)(233,0.3408)(234,0.3406)(235,0.3556)(236,0.3384)(237,0.3384)(238,0.3384)(239,0.3384)(240,0.3384)(241,0.3384)(242,0.3384)(243,0.3384)(244,0.3384)(245,0.3384)(246,0.3384)(247,0.3384)(248,0.3384)(249,0.3384)(250,0.3384)(251,0.3384)(252,0.3384)(253,0.3384)(254,0.3384)(255,0.3384)(256,0.3384)(257,0.3384)(258,0.3384)(259,0.3384)(260,0.3384)(261,0.3384)(262,0.3384)(263,0.3384)(264,0.3384)(265,0.3384)(266,0.3384)(267,0.3384)(268,0.3384)(269,0.3384)(270,0.3384)(271,0.3384)(272,0.3384)(273,0.3384)(274,0.3384)(275,0.3384)(276,0.3384)(277,0.3384)(278,0.3384)(279,0.3384)(280,0.3384)(281,0.3384)(282,0.3384)(283,0.3384)(284,0.3384)(285,0.3384)(286,0.3384)(287,0.3384)(288,0.3384)(289,0.3384)(290,0.3384)(291,0.3384)(292,0.3384)(293,0.3384)(294,0.3384)(295,0.3384)(296,0.3384)(297,0.3384)(298,0.3384)(299,0.3384)

    };  

\addplot[
    color=blue,
    line width=1pt,
    ]
    coordinates {
    (0,0.5264)(1,0.4248)(2,0.3871)(3,0.3751)(4,0.3665)(5,0.3526)(6,0.3802)(7,0.3743)(8,0.3652)(9,0.3638)(10,0.3514)(11,0.3618)(12,0.3435)(13,0.3348)(14,0.3553)(15,0.3331)(16,0.3301)(17,0.3274)(18,0.3232)(19,0.3379)(20,0.3210)(21,0.3223)(22,0.3205)(23,0.3275)(24,0.3189)(25,0.3300)(26,0.3257)(27,0.3132)(28,0.3150)(29,0.3194)(30,0.3152)(31,0.3169)(32,0.3167)(33,0.3302)(34,0.3290)(35,0.3134)(36,0.3201)(37,0.3186)(38,0.3154)(39,0.3404)(40,0.3147)(41,0.3349)(42,0.3116)(43,0.3097)(44,0.3317)(45,0.3208)(46,0.3250)(47,0.3111)(48,0.3192)(49,0.3140)(50,0.3310)(51,0.3114)(52,0.3099)(53,0.3339)(54,0.3281)(55,0.3222)(56,0.3206)(57,0.3212)(58,0.3288)(59,0.3078)(60,0.3156)(61,0.3071)(62,0.3143)(63,0.3173)(64,0.3086)(65,0.3164)(66,0.3095)(67,0.3072)(68,0.3056)(69,0.3077)(70,0.3193)(71,0.3093)(72,0.3079)(73,0.3159)(74,0.3091)(75,0.3105)(76,0.3102)(77,0.3209)(78,0.3094)(79,0.3125)(80,0.3065)(81,0.3291)(82,0.3062)(83,0.3104)(84,0.3157)(85,0.3143)(86,0.3127)(87,0.3064)(88,0.3140)(89,0.3105)(90,0.3071)(91,0.3121)(92,0.3047)(93,0.3041)(94,0.3055)(95,0.3172)(96,0.3052)(97,0.3066)(98,0.3129)(99,0.3059)(100,0.3096)(101,0.3055)(102,0.3067)(103,0.3277)(104,0.3070)(105,0.3104)(106,0.3065)(107,0.3104)(108,0.3078)(109,0.3053)(110,0.3056)(111,0.3150)(112,0.3059)(113,0.3135)(114,0.3063)(115,0.3107)(116,0.3074)(117,0.3062)(118,0.3076)(119,0.3160)(120,0.3101)(121,0.3145)(122,0.3094)(123,0.3059)(124,0.3175)(125,0.3057)(126,0.3054)(127,0.3088)(128,0.3056)(129,0.3243)(130,0.3270)(131,0.3126)(132,0.3059)(133,0.3081)(134,0.3041)(135,0.3080)(136,0.3065)(137,0.3072)(138,0.3117)(139,0.3143)(140,0.3116)(141,0.3112)(142,0.3150)(143,0.3057)(144,0.3145)(145,0.3188)(146,0.3090)(147,0.3102)(148,0.3084)(149,0.3165)(150,0.3127)(151,0.3093)(152,0.3089)(153,0.3065)(154,0.3068)(155,0.3266)(156,0.3086)(157,0.3156)(158,0.3086)(159,0.3210)(160,0.3098)(161,0.3093)(162,0.3040)(163,0.3132)(164,0.3272)(165,0.3294)(166,0.3190)(167,0.3332)(168,0.3057)(169,0.3103)(170,0.3081)(171,0.3040)(172,0.3150)(173,0.3079)(174,0.3050)(175,0.3083)(176,0.3079)(177,0.3125)(178,0.3253)(179,0.3096)(180,0.3054)(181,0.3208)(182,0.3234)(183,0.3074)(184,0.3202)(185,0.3220)(186,0.3184)(187,0.3134)(188,0.3201)(189,0.3074)(190,0.3053)(191,0.3364)(192,0.3068)(193,0.3030)(194,0.3030)(195,0.3030)(196,0.3030)(197,0.3030)(198,0.3030)(199,0.3030)(200,0.3030)(201,0.3030)(202,0.3030)(203,0.3030)(204,0.3030)(205,0.3030)(206,0.3030)(207,0.3030)(208,0.3030)(209,0.3030)(210,0.3030)(211,0.3030)(212,0.3030)(213,0.3030)(214,0.3030)(215,0.3030)(216,0.3030)(217,0.3030)(218,0.3030)(219,0.3030)(220,0.3030)(221,0.3030)(222,0.3030)(223,0.3030)(224,0.3030)(225,0.3030)(226,0.3030)(227,0.3030)(228,0.3030)(229,0.3030)(230,0.3030)(231,0.3030)(232,0.3030)(233,0.3030)(234,0.3030)(235,0.3030)(236,0.3030)(237,0.3030)(238,0.3030)(239,0.3030)(240,0.3030)(241,0.3030)(242,0.3030)(243,0.3030)(244,0.3030)(245,0.3030)(246,0.3030)(247,0.3030)(248,0.3030)(249,0.3030)(250,0.3030)(251,0.3030)(252,0.3030)(253,0.3030)(254,0.3030)(255,0.3030)(256,0.3030)(257,0.3030)(258,0.3030)(259,0.3030)(260,0.3030)(261,0.3030)(262,0.3030)(263,0.3030)(264,0.3030)(265,0.3030)(266,0.3030)(267,0.3030)(268,0.3030)(269,0.3030)(270,0.3030)(271,0.3030)(272,0.3030)(273,0.3030)(274,0.3030)(275,0.3030)(276,0.3030)(277,0.3030)(278,0.3030)(279,0.3030)(280,0.3030)(281,0.3030)(282,0.3030)(283,0.3030)(284,0.3030)(285,0.3030)(286,0.3030)(287,0.3030)(288,0.3030)(289,0.3030)(290,0.3030)(291,0.3030)(292,0.3030)(293,0.3030)(294,0.3030)(295,0.3030)(296,0.3030)(297,0.3030)(298,0.3030)(299,0.3030)
    };  

\addplot[
    color=red,
      line width=1pt,
    ]
    coordinates {
   (0,0.6032)(1,0.5467)(2,0.5112)(3,0.4814)(4,0.4622)(5,0.4512)(6,0.4436)(7,0.4281)(8,0.4184)(9,0.4169)(10,0.4056)(11,0.3984)(12,0.3909)(13,0.3854)(14,0.3847)(15,0.3742)(16,0.3732)(17,0.3684)(18,0.3666)(19,0.3591)(20,0.3554)(21,0.3519)(22,0.3476)(23,0.3482)(24,0.3445)(25,0.3432)(26,0.3449)(27,0.3370)(28,0.3380)(29,0.3345)(30,0.3340)(31,0.3308)(32,0.3337)(33,0.3293)(34,0.3300)(35,0.3263)(36,0.3276)(37,0.3257)(38,0.3325)(39,0.3245)(40,0.3273)(41,0.3291)(42,0.3228)(43,0.3176)(44,0.3191)(45,0.3172)(46,0.3191)(47,0.3215)(48,0.3151)(49,0.3177)(50,0.3183)(51,0.3158)(52,0.3209)(53,0.3210)(54,0.3163)(55,0.3158)(56,0.3144)(57,0.3147)(58,0.3135)(59,0.3172)(60,0.3198)(61,0.3142)(62,0.3182)(63,0.3146)(64,0.3103)(65,0.3182)(66,0.3121)(67,0.3165)(68,0.3139)(69,0.3147)(70,0.3219)(71,0.3104)(72,0.3144)(73,0.3136)(74,0.3128)(75,0.3192)(76,0.3139)(77,0.3090)(78,0.3108)(79,0.3136)(80,0.3181)(81,0.3158)(82,0.3095)(83,0.3196)(84,0.3136)(85,0.3139)(86,0.3092)(87,0.3081)(88,0.3146)(89,0.3092)(90,0.3138)(91,0.3106)(92,0.3105)(93,0.3100)(94,0.3212)(95,0.3100)(96,0.3206)(97,0.3084)(98,0.3130)(99,0.3064)(100,0.3097)(101,0.3080)(102,0.3124)(103,0.3102)(104,0.3091)(105,0.3092)(106,0.3129)(107,0.3090)(108,0.3067)(109,0.3055)(110,0.3075)(111,0.3112)(112,0.3067)(113,0.3089)(114,0.3060)(115,0.3104)(116,0.3128)(117,0.3111)(118,0.3064)(119,0.3112)(120,0.3076)(121,0.3080)(122,0.3057)(123,0.3088)(124,0.3095)(125,0.3073)(126,0.3085)(127,0.3104)(128,0.3127)(129,0.3084)(130,0.3053)(131,0.3140)(132,0.3106)(133,0.3061)(134,0.3070)(135,0.3071)(136,0.3051)(137,0.3032)(138,0.3051)(139,0.3093)(140,0.3040)(141,0.3063)(142,0.3029)(143,0.3067)(144,0.3077)(145,0.3035)(146,0.3079)(147,0.3060)(148,0.3052)(149,0.3066)(150,0.3048)(151,0.3044)(152,0.3057)(153,0.3051)(154,0.3044)(155,0.3094)(156,0.3092)(157,0.3072)(158,0.3138)(159,0.3053)(160,0.3092)(161,0.3047)(162,0.3047)(163,0.3042)(164,0.3057)(165,0.3082)(166,0.3026)(167,0.3073)(168,0.3069)(169,0.3086)(170,0.3056)(171,0.3062)(172,0.3115)(173,0.3129)(174,0.3083)(175,0.3080)(176,0.3044)(177,0.3032)(178,0.3072)(179,0.3053)(180,0.3188)(181,0.3074)(182,0.3026)(183,0.3032)(184,0.3062)(185,0.3021)(186,0.3048)(187,0.3066)(188,0.3025)(189,0.3039)(190,0.3031)(191,0.3025)(192,0.3048)(193,0.3044)(194,0.3084)(195,0.3055)(196,0.3031)(197,0.3016)(198,0.3053)(199,0.3053)(200,0.3029)(201,0.3074)(202,0.3087)(203,0.3059)(204,0.3028)(205,0.3057)(206,0.3045)(207,0.3047)(208,0.3061)(209,0.3073)(210,0.3057)(211,0.3032)(212,0.3039)(213,0.3032)(214,0.3039)(215,0.3038)(216,0.3038)(217,0.3011)(218,0.3024)(219,0.3050)(220,0.3034)(221,0.3015)(222,0.3061)(223,0.3023)(224,0.3051)(225,0.3044)(226,0.3113)(227,0.3039)(228,0.3087)(229,0.3028)(230,0.3051)(231,0.3083)(232,0.3055)(233,0.3022)(234,0.3023)(235,0.3072)(236,0.3107)(237,0.3036)(238,0.3049)(239,0.3038)(240,0.3032)(241,0.3066)(242,0.3061)(243,0.3057)(244,0.3024)(245,0.3049)(246,0.3027)(247,0.3096)(248,0.3098)(249,0.3042)(250,0.3022)(251,0.3071)(252,0.3027)(253,0.3082)(254,0.3056)(255,0.3060)(256,0.3059)(257,0.3067)(258,0.3085)(259,0.3029)(260,0.3048)(261,0.3062)(262,0.3039)(263,0.3010)(264,0.3052)(265,0.3092)(266,0.3014)(267,0.3009)(268,0.3123)(269,0.3109)(270,0.3030)(271,0.3019)(272,0.3058)(273,0.3014)(274,0.3021)(275,0.3015)(276,0.3022)(277,0.3057)(278,0.3004)(279,0.3029)(280,0.3015)(281,0.3029)(282,0.3023)(283,0.3046)(284,0.3009)(285,0.3027)(286,0.3054)(287,0.2999)(288,0.2999)(289,0.2999)(290,0.2999)(291,0.2999)(292,0.2999)(293,0.2999)(294,0.2999)(295,0.2999)(296,0.2999)(297,0.2999)(298,0.2999)(299,0.2999)
    };  

\end{axis}

\hskip 120pt

\begin{axis}[
    title={Movielens},
    xlabel={Epoch},
    ylabel={RMSE},
    xmin=0, xmax=300,
    ymin=0.42, ymax=0.55,
    xtick={0,100,200,300},
    ytick={0.42, 0.45, 0.5, 0.55},
    ymajorgrids=false,
    xmajorgrids=false,
    grid style=dashed,
    legend entries={FM, GEM(type:b), GEM(type:a)},
    legend style={
            font=\tiny,
            /tikz/every even column/.append style={column sep=0.5cm}}
]
\addplot[
    color=black,
    ]
    coordinates {
  (0,0.6239)(1,0.5345)(2,0.5127)(3,0.4993)(4,0.4926)(5,0.4870)(6,0.4827)(7,0.4829)(8,0.4797)(9,0.4805)(10,0.4792)(11,0.4748)(12,0.4759)(13,0.4747)(14,0.4749)(15,0.4740)(16,0.4701)(17,0.4718)(18,0.4683)(19,0.4694)(20,0.4687)(21,0.4684)(22,0.4684)(23,0.4676)(24,0.4656)(25,0.4657)(26,0.4642)(27,0.4629)(28,0.4665)(29,0.4624)(30,0.4623)(31,0.4680)(32,0.4594)(33,0.4633)(34,0.4618)(35,0.4603)(36,0.4598)(37,0.4614)(38,0.4593)(39,0.4579)(40,0.4621)(41,0.4641)(42,0.4624)(43,0.4593)(44,0.4583)(45,0.4584)(46,0.4596)(47,0.4563)(48,0.4570)(49,0.4613)(50,0.4544)(51,0.4577)(52,0.4542)(53,0.4555)(54,0.4551)(55,0.4556)(56,0.4566)(57,0.4570)(58,0.4568)(59,0.4552)(60,0.4537)(61,0.4549)(62,0.4534)(63,0.4584)(64,0.4554)(65,0.4568)(66,0.4596)(67,0.4565)(68,0.4549)(69,0.4549)(70,0.4531)(71,0.4543)(72,0.4544)(73,0.4525)(74,0.4522)(75,0.4549)(76,0.4530)(77,0.4528)(78,0.4529)(79,0.4526)(80,0.4546)(81,0.4534)(82,0.4517)(83,0.4495)(84,0.4528)(85,0.4557)(86,0.4527)(87,0.4546)(88,0.4515)(89,0.4525)(90,0.4532)(91,0.4544)(92,0.4540)(93,0.4502)(94,0.4517)(95,0.4522)(96,0.4497)(97,0.4493)(98,0.4537)(99,0.4522)(100,0.4514)(101,0.4479)(102,0.4516)(103,0.4521)(104,0.4476)(105,0.4499)(106,0.4526)(107,0.4522)(108,0.4524)(109,0.4534)(110,0.4576)(111,0.4500)(112,0.4528)(113,0.4508)(114,0.4514)(115,0.4555)(116,0.4549)(117,0.4504)(118,0.4510)(119,0.4519)(120,0.4550)(121,0.4491)(122,0.4501)(123,0.4506)(124,0.4494)(125,0.4533)(126,0.4493)(127,0.4530)(128,0.4484)(129,0.4492)(130,0.4517)(131,0.4520)(132,0.4514)(133,0.4516)(134,0.4491)(135,0.4496)(136,0.4510)(137,0.4528)(138,0.4527)(139,0.4503)(140,0.4493)(141,0.4473)(142,0.4499)(143,0.4518)(144,0.4505)(145,0.4511)(146,0.4498)(147,0.4477)(148,0.4518)(149,0.4502)(150,0.4502)(151,0.4479)(152,0.4530)(153,0.4490)(154,0.4495)(155,0.4484)(156,0.4482)(157,0.4509)(158,0.4513)(159,0.4524)(160,0.4518)(161,0.4508)(162,0.4495)(163,0.4485)(164,0.4513)(165,0.4506)(166,0.4511)(167,0.4559)(168,0.4495)(169,0.4494)(170,0.4489)(171,0.4533)(172,0.4504)(173,0.4525)(174,0.4492)(175,0.4496)(176,0.4503)(177,0.4485)(178,0.4479)(179,0.4502)(180,0.4490)(181,0.4528)(182,0.4543)(183,0.4512)(184,0.4509)(185,0.4469)(186,0.4474)(187,0.4474)(188,0.4515)(189,0.4491)(190,0.4487)(191,0.4509)(192,0.4492)(193,0.4527)(194,0.4492)(195,0.4509)(196,0.4502)(197,0.4494)(198,0.4472)(199,0.4474)(200,0.4480)(201,0.4474)(202,0.4500)(203,0.4488)(204,0.4489)(205,0.4478)(206,0.4461)(207,0.4504)(208,0.4473)(209,0.4471)(210,0.4496)(211,0.4473)(212,0.4492)(213,0.4486)(214,0.4463)(215,0.4491)(216,0.4494)(217,0.4457)(218,0.4478)(219,0.4472)(220,0.4484)(221,0.4456)(222,0.4472)(223,0.4496)(224,0.4481)(225,0.4483)(226,0.4506)(227,0.4492)(228,0.4488)(229,0.4509)(230,0.4482)(231,0.4509)(232,0.4479)(233,0.4510)(234,0.4467)(235,0.4484)(236,0.4478)(237,0.4475)(238,0.4484)(239,0.4460)(240,0.4505)(241,0.4465)(242,0.4492)(243,0.4500)(244,0.4476)(245,0.4487)(246,0.4517)(247,0.4490)(248,0.4473)(249,0.4472)(250,0.4489)(251,0.4495)(252,0.4467)(253,0.4479)(254,0.4472)(255,0.4506)(256,0.4473)(257,0.4463)(258,0.4518)(259,0.4498)(260,0.4506)(261,0.4489)(262,0.4479)(263,0.4470)(264,0.4486)(265,0.4500)(266,0.4517)(267,0.4482)(268,0.4505)(269,0.4473)(270,0.4483)(271,0.4482)(272,0.4450)(273,0.4450)(274,0.4450)(275,0.4450)(276,0.4450)(277,0.4450)(278,0.4450)(279,0.4450)(280,0.4450)(281,0.4450)(282,0.4450)(283,0.4450)(284,0.4450)(285,0.4450)(286,0.4450)(287,0.4450)(288,0.4450)(289,0.4450)(290,0.4450)(291,0.4450)(292,0.4450)(293,0.4450)(294,0.4450)(295,0.4450)(296,0.4450)(297,0.4450)(298,0.4450)(299,0.4450)
    }; 

\addplot[
    color=blue,
    ]
    coordinates {
     (0,0.5992)(1,0.5376)(2,0.5189)(3,0.5084)(4,0.4958)(5,0.4853)(6,0.4816)(7,0.4770)(8,0.4746)(9,0.4697)(10,0.4680)(11,0.4675)(12,0.4685)(13,0.4636)(14,0.4586)(15,0.4611)(16,0.4598)(17,0.4603)(18,0.4580)(19,0.4592)(20,0.4571)(21,0.4553)(22,0.4571)(23,0.4565)(24,0.4565)(25,0.4498)(26,0.4528)(27,0.4552)(28,0.4488)(29,0.4509)(30,0.4472)(31,0.4480)(32,0.4517)(33,0.4455)(34,0.4515)(35,0.4465)(36,0.4507)(37,0.4486)(38,0.4487)(39,0.4459)(40,0.4427)(41,0.4677)(42,0.4419)(43,0.4418)(44,0.4430)(45,0.4476)(46,0.4414)(47,0.4489)(48,0.4426)(49,0.4399)(50,0.4421)(51,0.4533)(52,0.4433)(53,0.4453)(54,0.4386)(55,0.4397)(56,0.4415)(57,0.4456)(58,0.4378)(59,0.4411)(60,0.4394)(61,0.4388)(62,0.4369)(63,0.4402)(64,0.4428)(65,0.4361)(66,0.4420)(67,0.4421)(68,0.4449)(69,0.4384)(70,0.4366)(71,0.4368)(72,0.4436)(73,0.4359)(74,0.4451)(75,0.4435)(76,0.4566)(77,0.4456)(78,0.4362)(79,0.4415)(80,0.4382)(81,0.4510)(82,0.4382)(83,0.4384)(84,0.4402)(85,0.4360)(86,0.4397)(87,0.4359)(88,0.4368)(89,0.4373)(90,0.4407)(91,0.4362)(92,0.4439)(93,0.4429)(94,0.4411)(95,0.4409)(96,0.4346)(97,0.4343)(98,0.4343)(99,0.4343)(100,0.4343)(101,0.4343)(102,0.4343)(103,0.4343)(104,0.4343)(105,0.4343)(106,0.4343)(107,0.4343)(108,0.4343)(109,0.4343)(110,0.4343)(111,0.4343)(112,0.4343)(113,0.4343)(114,0.4343)(115,0.4343)(116,0.4343)(117,0.4343)(118,0.4343)(119,0.4343)(120,0.4343)(121,0.4343)(122,0.4343)(123,0.4343)(124,0.4343)(125,0.4343)(126,0.4343)(127,0.4343)(128,0.4343)(129,0.4343)(130,0.4343)(131,0.4343)(132,0.4343)(133,0.4343)(134,0.4343)(135,0.4343)(136,0.4343)(137,0.4343)(138,0.4343)(139,0.4343)(140,0.4343)(141,0.4343)(142,0.4343)(143,0.4343)(144,0.4343)(145,0.4343)(146,0.4343)(147,0.4343)(148,0.4343)(149,0.4343)(150,0.4343)(151,0.4343)(152,0.4343)(153,0.4343)(154,0.4343)(155,0.4343)(156,0.4343)(157,0.4343)(158,0.4343)(159,0.4343)(160,0.4343)(161,0.4343)(162,0.4343)(163,0.4343)(164,0.4343)(165,0.4343)(166,0.4343)(167,0.4343)(168,0.4343)(169,0.4343)(170,0.4343)(171,0.4343)(172,0.4343)(173,0.4343)(174,0.4343)(175,0.4343)(176,0.4343)(177,0.4343)(178,0.4343)(179,0.4343)(180,0.4343)(181,0.4343)(182,0.4343)(183,0.4343)(184,0.4343)(185,0.4343)(186,0.4343)(187,0.4343)(188,0.4343)(189,0.4343)(190,0.4343)(191,0.4343)(192,0.4343)(193,0.4343)(194,0.4343)(195,0.4343)(196,0.4343)(197,0.4343)(198,0.4343)(199,0.4343)(200,0.4343)(201,0.4343)(202,0.4343)(203,0.4343)(204,0.4343)(205,0.4343)(206,0.4343)(207,0.4343)(208,0.4343)(209,0.4343)(210,0.4343)(211,0.4343)(212,0.4343)(213,0.4343)(214,0.4343)(215,0.4343)(216,0.4343)(217,0.4343)(218,0.4343)(219,0.4343)(220,0.4343)(221,0.4343)(222,0.4343)(223,0.4343)(224,0.4343)(225,0.4343)(226,0.4343)(227,0.4343)(228,0.4343)(229,0.4343)(230,0.4343)(231,0.4343)(232,0.4343)(233,0.4343)(234,0.4343)(235,0.4343)(236,0.4343)(237,0.4343)(238,0.4343)(239,0.4343)(240,0.4343)(241,0.4343)(242,0.4343)(243,0.4343)(244,0.4343)(245,0.4343)(246,0.4343)(247,0.4343)(248,0.4343)(249,0.4343)(250,0.4343)(251,0.4343)(252,0.4343)(253,0.4343)(254,0.4343)(255,0.4343)(256,0.4343)(257,0.4343)(258,0.4343)(259,0.4343)(260,0.4343)(261,0.4343)(262,0.4343)(263,0.4343)(264,0.4343)(265,0.4343)(266,0.4343)(267,0.4343)(268,0.4343)(269,0.4343)(270,0.4343)(271,0.4343)(272,0.4343)(273,0.4343)(274,0.4343)(275,0.4343)(276,0.4343)(277,0.4343)(278,0.4343)(279,0.4343)(280,0.4343)(281,0.4343)(282,0.4343)(283,0.4343)(284,0.4343)(285,0.4343)(286,0.4343)(287,0.4343)(288,0.4343)(289,0.4343)(290,0.4343)(291,0.4343)(292,0.4343)(293,0.4343)(294,0.4343)(295,0.4343)(296,0.4343)(297,0.4343)(298,0.4343)(299,0.4343)
    }; 

\addplot[
    color=red,
     line width=1pt,
    ]
    coordinates {
       (0,0.5903)(1,0.5548)(2,0.5290)(3,0.5129)(4,0.4987)(5,0.4910)(6,0.4822)(7,0.4766)(8,0.4746)(9,0.4692)(10,0.4654)(11,0.4626)(12,0.4597)(13,0.4592)(14,0.4577)(15,0.4559)(16,0.4544)(17,0.4549)(18,0.4536)(19,0.4524)(20,0.4517)(21,0.4511)(22,0.4524)(23,0.4499)(24,0.4485)(25,0.4494)(26,0.4490)(27,0.4479)(28,0.4472)(29,0.4465)(30,0.4461)(31,0.4468)(32,0.4460)(33,0.4444)(34,0.4463)(35,0.4446)(36,0.4447)(37,0.4433)(38,0.4456)(39,0.4432)(40,0.4441)(41,0.4428)(42,0.4427)(43,0.4423)(44,0.4436)(45,0.4425)(46,0.4414)(47,0.4427)(48,0.4414)(49,0.4419)(50,0.4420)(51,0.4411)(52,0.4418)(53,0.4405)(54,0.4412)(55,0.4396)(56,0.4394)(57,0.4407)(58,0.4399)(59,0.4419)(60,0.4399)(61,0.4393)(62,0.4392)(63,0.4392)(64,0.4390)(65,0.4401)(66,0.4384)(67,0.4387)(68,0.4386)(69,0.4383)(70,0.4384)(71,0.4382)(72,0.4390)(73,0.4385)(74,0.4382)(75,0.4380)(76,0.4371)(77,0.4368)(78,0.4383)(79,0.4372)(80,0.4370)(81,0.4369)(82,0.4374)(83,0.4378)(84,0.4372)(85,0.4363)(86,0.4368)(87,0.4383)(88,0.4384)(89,0.4365)(90,0.4366)(91,0.4361)(92,0.4365)(93,0.4358)(94,0.4362)(95,0.4353)(96,0.4354)(97,0.4354)(98,0.4355)(99,0.4373)(100,0.4355)(101,0.4365)(102,0.4354)(103,0.4354)(104,0.4354)(105,0.4358)(106,0.4350)(107,0.4359)(108,0.4345)(109,0.4351)(110,0.4354)(111,0.4341)(112,0.4343)(113,0.4340)(114,0.4346)(115,0.4347)(116,0.4341)(117,0.4341)(118,0.4341)(119,0.4344)(120,0.4344)(121,0.4343)(122,0.4339)(123,0.4341)(124,0.4339)(125,0.4352)(126,0.4342)(127,0.4350)(128,0.4331)(129,0.4331)(130,0.4331)(131,0.4331)(132,0.4331)(133,0.4331)(134,0.4331)(135,0.4331)(136,0.4331)(137,0.4331)(138,0.4331)(139,0.4331)(140,0.4331)(141,0.4331)(142,0.4331)(143,0.4331)(144,0.4331)(145,0.4331)(146,0.4331)(147,0.4331)(148,0.4331)(149,0.4331)(150,0.4331)(151,0.4331)(152,0.4331)(153,0.4331)(154,0.4331)(155,0.4331)(156,0.4331)(157,0.4331)(158,0.4331)(159,0.4331)(160,0.4331)(161,0.4331)(162,0.4331)(163,0.4331)(164,0.4331)(165,0.4331)(166,0.4331)(167,0.4331)(168,0.4331)(169,0.4331)(170,0.4331)(171,0.4331)(172,0.4331)(173,0.4331)(174,0.4331)(175,0.4331)(176,0.4331)(177,0.4331)(178,0.4331)(179,0.4331)(180,0.4331)(181,0.4331)(182,0.4331)(183,0.4331)(184,0.4331)(185,0.4331)(186,0.4331)(187,0.4331)(188,0.4331)(189,0.4331)(190,0.4331)(191,0.4331)(192,0.4331)(193,0.4331)(194,0.4331)(195,0.4331)(196,0.4331)(197,0.4331)(198,0.4331)(199,0.4331)(200,0.4331)(201,0.4331)(202,0.4331)(203,0.4331)(204,0.4331)(205,0.4331)(206,0.4331)(207,0.4331)(208,0.4331)(209,0.4331)(210,0.4331)(211,0.4331)(212,0.4331)(213,0.4331)(214,0.4331)(215,0.4331)(216,0.4331)(217,0.4331)(218,0.4331)(219,0.4331)(220,0.4331)(221,0.4331)(222,0.4331)(223,0.4331)(224,0.4331)(225,0.4331)(226,0.4331)(227,0.4331)(228,0.4331)(229,0.4331)(230,0.4331)(231,0.4331)(232,0.4331)(233,0.4331)(234,0.4331)(235,0.4331)(236,0.4331)(237,0.4331)(238,0.4331)(239,0.4331)(240,0.4331)(241,0.4331)(242,0.4331)(243,0.4331)(244,0.4331)(245,0.4331)(246,0.4331)(247,0.4331)(248,0.4331)(249,0.4331)(250,0.4331)(251,0.4331)(252,0.4331)(253,0.4331)(254,0.4331)(255,0.4331)(256,0.4331)(257,0.4331)(258,0.4331)(259,0.4331)(260,0.4331)(261,0.4331)(262,0.4331)(263,0.4331)(264,0.4331)(265,0.4331)(266,0.4331)(267,0.4331)(268,0.4331)(269,0.4331)(270,0.4331)(271,0.4331)(272,0.4331)(273,0.4331)(274,0.4331)(275,0.4331)(276,0.4331)(277,0.4331)(278,0.4331)(279,0.4331)(280,0.4331)(281,0.4331)(282,0.4331)(283,0.4331)(284,0.4331)(285,0.4331)(286,0.4331)(287,0.4331)(288,0.4331)(289,0.4331)(290,0.4331)(291,0.4331)(292,0.4331)(293,0.4331)(294,0.4331)(295,0.4331)(296,0.4331)(297,0.4331)(298,0.4331)(299,0.4331)
    };   

\end{axis}
\end{tikzpicture}
\caption{Impact of the graph construction}
\label{fig:nodetype}
\end{figure}

\subsubsection{The Impact of Neighbor Sampling Ratio}
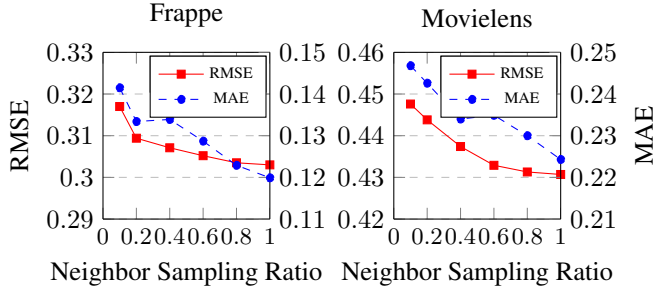
\begin{figure}[tb]
\begin{tikzpicture}
\pgfplotsset{width=3.8cm, height=3.8cm, compat=1.5}
\begin{axis}[
    axis y line*=left,
    title={Frappe},
    xlabel={Neighbor Sampling Ratio},
    ylabel={RMSE},
    xmin=0.0, xmax=1.0,
    ymin=0.29, ymax=0.33,
    xtick={0.0,0.2,0.4,0.6,0.8,1.0},
    ytick={0.29,0.30,0.31,0.32,0.33},
    xticklabel style={font=\small},
    ymajorgrids=true,
    xmajorgrids=false,
    grid style=dashed,
      legend style={font=\tiny,
     /tikz/every even column/.append style={column sep=0.5cm} }
]
\addplot[
    color=red,
    mark = square*,
    mark size=1.5pt,
    ]
  coordinates{
    (0.1, 0.3170) (0.2, 0.3094) (0.4, 0.3071)
    (0.6, 0.3052) (0.8, 0.3035) (1.0, 0.3030)
    };
\label{plot_frappe_rmse}
\addlegendentry{RMSE}
\end{axis}

\begin{axis}[
  axis y line*=right,
  xmin=0.0, xmax=1.0,
  xtick={0.0,0.2,0.4,0.6,0.8,1.0},
  xticklabel style={font=\small},
  axis x line=none,
  ymin=0.11, ymax=0.15,
  ytick={0.11,0.12,0.13,0.14,0.15},
  ymajorgrids=false,
  xmajorgrids=false,
  grid style=dashed,
    legend style={font=\tiny,
     /tikz/every even column/.append style={column sep=0.5cm} }
]
\addlegendimage{/pgfplots/refstyle=plot_frappe_rmse}
\addlegendentry{RMSE}
\addplot[mark=*,blue,dashed = true, mark size=1.5pt,]
  coordinates{
  (0.1, 0.1415) (0.2, 0.1334) (0.4, 0.1339)
  (0.6,0.1287) (0.8, 0.1229) (1.0, 0.1199)
};
\addlegendentry{MAE}
\end{axis}

\hskip 110pt

\begin{axis}[
    title={Movielens},
    axis y line*=left,
    xlabel={Neighbor Sampling Ratio},
    xmin=0.0, xmax=1.0,
    ymin=0.42, ymax=0.46,
    xticklabel style={font=\small},
    xtick={0.0,0.2,0.4,0.6,0.8,1.0},
    ytick={0.42,0.43,0.44,0.45,0.46},
    ymajorgrids=true,
    xmajorgrids=false,
    grid style=dashed,
    legend style={font=\tiny,
     /tikz/every even column/.append style={column sep=0.5cm} }
]
\addplot[
    color=red,
    mark = square*,
    mark size=1.5pt,
    ]
  coordinates{
     (0.1, 0.4476) (0.2, 0.4438) (0.4, 0.4374)
     (0.6, 0.4329) (0.8, 0.4313) (1.0, 0.4307)
};
\label{plot_ml_rmse}
\addlegendentry{RMSE}
\end{axis}

\begin{axis}[
  axis y line*=right,
  axis x line=none,
  ylabel={MAE},
  xmin=0.0, xmax=1.0,
  ymin=0.21, ymax=0.25,
  xtick={0.0,0.2,0.4,0.6,0.8,1.0},
  ytick={0.21,0.22, 0.23, 0.24, 0.25},
  xticklabel style={font=\small},
  ymajorgrids=false,
  xmajorgrids=false,
  grid style=dashed,
  legend style={font=\tiny,
     /tikz/every even column/.append style={column sep=0.5cm} }
]
\addlegendimage{/pgfplots/refstyle=plot_ml_rmse}
\addlegendentry{RMSE}
\addplot[mark=*,blue,dashed = true, mark size=1.5pt,]
  coordinates{
     (0.1, 0.2468) (0.2, 0.2426) (0.4, 0.2340)
     (0.6, 0.2349) (0.8, 0.2300) (1.0, 0.2243)
};
\addlegendentry{MAE}
\label{plot_ml_rmse1}
\end{axis}

\end{tikzpicture}
\caption{Impact of neighbor sampling ratio}
\label{fig:sampling}
\end{figure}

In Figure \ref{fig:sampling}, we show the result of sampling different proportions of the neighbors on the graph. In other words, the high-order embedding of feature is influenced by how many related features. On the two evaluation metrics and both datasets, we can reach the same conclusion that with the increase of sampling ratio, the model performance keeps improving. We can have a straightforward explanation that when the sampling ratio is $0$, we actually don't encode any high-order information into the embedding and our model downgrade to the base model. With the increase of sampling ratio, the embedding starts to encode more and more diverse high-order signals and thus improves the model performance.

\subsection{(RQ3) Hyper-parameter Study}
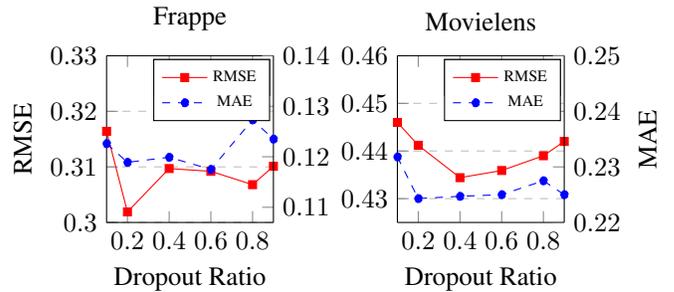
\begin{figure}[tb]
\begin{tikzpicture}
\pgfplotsset{width=3.8cm, height=3.8cm, compat=1.5}
\begin{axis}[
    axis y line*=left,
    title={Frappe},
    ylabel={RMSE},
    xlabel={Dropout Ratio},
    xmin=0.1, xmax=0.9,
    ymin=0.30, ymax=0.33,
    xtick={0.2,0.4,0.6,0.8},
    ytick={0.30,0.31,0.32,0.33},
    ymajorgrids=true,
    xmajorgrids=false,
    grid style=dashed,
      legend style={font=\tiny,
     /tikz/every even column/.append style={column sep=0.5cm} }
]
\addplot[
    color=red,
    mark = square*,
    mark size=1.5pt,
    ]
  coordinates{
    (0.1, 0.3164) (0.2, 0.3019) (0.4, 0.3097)
    (0.6, 0.3092) (0.8, 0.3068) (0.9, 0.3101)
};
\label{plot_frappe_rmse}
\addlegendentry{RMSE}
\end{axis}

\begin{axis}[
  axis y line*=right,
  xmin=0.1, xmax=0.9,
  xtick={0.2,0.4,0.6,0.8},
  axis x line=none,
  ymin=0.107, ymax=0.14,
  ytick={0.10,0.11,0.12,0.13,0.14},
  ymajorgrids=false,
  xmajorgrids=false,
  grid style=dashed,
    legend style={font=\tiny,
     /tikz/every even column/.append style={column sep=0.5cm} }
]
\addlegendimage{/pgfplots/refstyle=plot_frappe_rmse}
\addlegendentry{RMSE}
\addplot[mark=*,blue, mark size=1.5pt,dashed = true]
  coordinates{
    (0.1,0.12261) (0.2, 0.1189) (0.4, 0.1199)
    (0.6, 0.1175) (0.8, 0.1273) (0.9, 0.1235)
};
\addlegendentry{MAE}
\end{axis}

\hskip 110pt

\begin{axis}[
    title={Movielens},
    axis y line*=left,
    xlabel={Dropout Ratio},
    xmin=0.1, xmax=0.9,
    ymin=0.425, ymax=0.46,
    xtick={0.0,0.2,0.4,0.6,0.8},
    ytick={0.42,0.43,0.44,0.45,0.46},
    ymajorgrids=true,
    xmajorgrids=false,
    grid style=dashed,
    legend style={font=\tiny,
     /tikz/every even column/.append style={column sep=0.5cm} }
]
\addplot[
    color=red,
    mark = square*,
    mark size=1.5pt,
    ]
  coordinates{
     (0.1, 0.4460) (0.2, 0.4412) (0.4, 0.4344)
     (0.6, 0.4359) (0.8, 0.4390) (0.9, 0.4420)
};
\label{plot_ml_rmse}
\addlegendentry{RMSE}
\end{axis}

\begin{axis}[
  axis y line*=right,
  xmin=0.1, xmax=0.9,
  ylabel={MAE},
  axis x line=none,
  ymin=0.22, ymax=0.25,
  ytick={0.22,0.23,0.24,0.25},
  ymajorgrids=false,
  xmajorgrids=false,
  grid style=dashed,
  legend style={font=\tiny,
     /tikz/every even column/.append style={column sep=0.5cm} }
]
\addlegendimage{/pgfplots/refstyle=plot_ml_rmse}\addlegendentry{RMSE}
\addplot[mark=*,blue, mark size=1.5pt,dashed = true]
  coordinates{
    (0.1, 0.2318) (0.2, 0.2243) (0.4, 0.2247)
    (0.6, 0.2250) (0.8,0.2275) (0.9, 0.2250)
};
\addlegendentry{MAE}
\end{axis}
\end{tikzpicture}
\caption{Impact of dropout ratio}
\label{fig:dropout}
\end{figure}

We use dropout in the GCN layer to prevent over-fitting of the model. In Figure \ref{fig:dropout}, we plot the performance change regarding to different dropout ratios. It can be seen that the model performance improves significantly when dropout was adjusted to a suitable value. Specifically, on the Frappe dataset, the best performance of RMSE and MAE is achieved when the dropout ratio is set as 0.2 and 0.6, respectively. On the Movielens dataset, the best performance of RMSE and MAE is achieved when the dropout ratio is set as $0.2$ and $0.4$.

\section{Conclusions and Future Work}
In this paper, different from conventional FM-based methods which attempt to capture high-order interaction signals by designing more complex interaction functions, we propose to encode high-order information from the embedding level. Based on such motivation, we propose the GEM approach which can be integrated with plenty of FM-based methods. It can be seen as the combination of cross features and the automatic interaction learning of FM-based methods. To make it more specific, in this paper we propose to utilize GCN to generate high-order embeddings. To verify the effectiveness of our approach, we integrated GEM with several state-of-the-art base models and conduct extensive experiments on two public accessible datasets. Experimental results demonstrate that incorporating GEM essentially improves the model performance. It also confirms our proposition that considering high-order information in the embedding level is necessary for better prediction.

Future work includes exploiting other techniques such as Transformer as the embedding function and more experiments about ranking-based tasks. Besides, we are also interested in designing adaptive negative samplers fro recommendation from the graph perspective.




\bibliographystyle{named}
\bibliography{ijcai20}

\end{document}